\documentclass{article}

\usepackage[preprint]{neurips_2026}

\usepackage[utf8]{inputenc}
\usepackage[T1]{fontenc}
\usepackage{hyperref}
\usepackage{url}
\usepackage{xurl}
\usepackage{booktabs}
\usepackage{amsfonts}
\usepackage{amsmath}
\usepackage{amssymb}
\usepackage{mathtools}
\usepackage{nicefrac}
\usepackage{microtype}
\usepackage{xcolor}
\usepackage{graphicx}
\usepackage{array}
\usepackage{multirow}
\usepackage{makecell}
\usepackage{threeparttable}
\usepackage{float}
\usepackage{placeins}
\usepackage{caption}
\usepackage{subcaption}
\usepackage{enumitem}
\usepackage{amsthm}
\newtheorem{proposition}{Proposition}

\hypersetup{hidelinks}
\definecolor{pandoBlue}{HTML}{1F4E79}

\graphicspath{{figs/}}

\newcommand{\ours}{PANDO}

\title{%
  \texorpdfstring{%
    \begin{tabular*}{0.96\textwidth}{@{}p{0.73\textwidth}@{\extracolsep{\fill}}c@{}}
      \raggedright PANDO: Efficient Multimodal AI Agents via Online Skill Distillation
      &
      \raisebox{-0.35\height}{\includegraphics[height=3.4em]{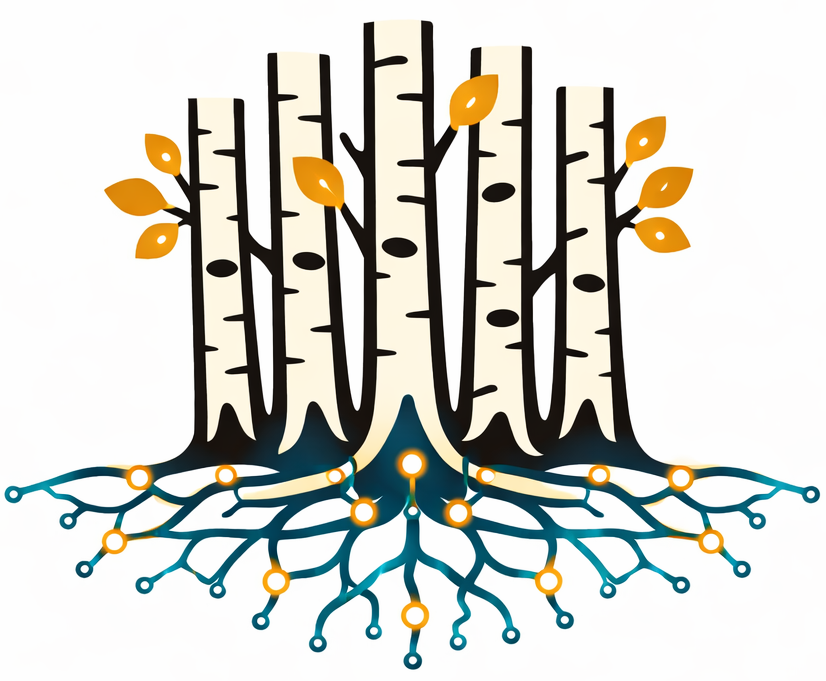}}
    \end{tabular*}%
  }{PANDO: Efficient Multimodal AI Agents via Online Skill Distillation}%
}

\author{
  Yubo Li \hspace{2em}
  Yidi Miao \hspace{2em}
  Yuntian Shen$^\dagger$ \hspace{2em}
  Yuxin Liu$^\dagger$ \\
  \texttt{\{yubol, yidim, yuntian2, yuxinli2\}@andrew.cmu.edu} \\
  $^\dagger$Co-third authors.
}

\begin{document}

\maketitle

\begin{abstract}
Recent multimodal web-agent gains have largely been bought through a simple token economy: spend more inference on rollout search, verifier passes, offline discovery, or specialist stacks. We ask whether an agent can instead become cheaper as it accumulates experience. A trajectory analysis on VisualWebArena identifies repeat-action loops, hidden discovery cost, and low prompt-cache reuse as recurring inefficiencies. We introduce \ours{}, a single-rollout online skill-distillation framework with a structured Skill Library, progress reflection, confidence-based demotion, hierarchical routing, visual compression, and cache-aware prompting. On all 910 VWA tasks, \ours{} reaches 58.3\% success, surpassing SGV (54.0\%) and our WALT reproduction (45.2\%) while using 58\% fewer tokens than SGV and 61\% fewer than WALT, with no pre-evaluation discovery budget. A 300-task ablation shows that rules and routines provide most of the success lift, whereas routing / compression / cache-aware prompting convert the larger library into lower marginal token load. We report Action Repetition Rate, Step Overhead Ratio, and Prompt Cache Utilization to make trajectory-level efficiency visible beyond terminal success.
\end{abstract}

\section{Introduction}

\begin{center}
{\footnotesize\textcolor{pandoBlue}{\emph{Many visible trunks, one shared root: Pando does not grow by restarting; it grows by remembering.}\footnotemark}}
\end{center}
\footnotetext{Appendix~\ref{app:naming} explains the name and its connection to the system design.}

The field has learned a remarkably effective recipe for better AI performance: spend more tokens. Larger contexts, longer chains of thought, self-consistency, verifier passes, tool-discovery phases, and best-of-$N$ rollouts all convert additional inference into higher benchmark scores. This creates a \emph{token economy} for agents: tokens are the currency used to buy accuracy, but they also determine marginal inference load, latency, cacheability, energy use, and the hidden liabilities of pre-evaluation discovery. That trade has been productive, but it is no longer a bookkeeping detail. Inference dominates the ML compute lifecycle~\citep{luccioni2024power}, production systems increasingly serve long reasoning traces~\citep{oviedo2025energy}, and data-center energy demand is becoming a first-order resource and environmental constraint~\citep{iea_energy_ai,lbnl_datacenter_2024}. The central question is therefore shifting from \emph{can we make the model better if we spend more?} to \emph{can we make the agent better without spending more every time?}

Computer-use agents make this question urgent. They are moving from demonstrations toward practical browser and desktop automation, but their operating mode is token hungry by construction: they process screenshots at every step, maintain long interaction histories, call planners and reflectors, and retry when grounding fails. Recent desktop-agent studies report 1.4--2.7$\times$ human step counts and 75--94\% of latency in planning / reflection~\citep{abhyankar2025osworld}. Frontier systems often push the same direction: behavior best-of-$N$ can multiply single-rollout compute by ten~\citep{gonzalez2025unreasonable}, while reasoning-heavy backbones inflate output-token budgets~\citep{oviedo2025energy}. Thus the token economics of computer-use agents are trajectory-level economics: the unit is not one prompt, but a stream of observations, plans, actions, reflections, and reusable or discarded experience.

We study this tension on VisualWebArena (VWA)~\citep{koh2024visualwebarena}. A trajectory audit of $1{,}000{+}$ baseline rollouts reveals three concrete sources of wasted work: \textbf{repeat-action loops} (34--42\% of image-annotated failures), \textbf{off-benchmark tool discovery} in systems such as WALT~\citep{prabhu2026walt}, and \textbf{prompt-architecture inefficiency}, where text / caption methods have prompt-cache utilization below 11\%. These are not generic ``model is weak'' errors; they are mechanistic inefficiencies that can be attacked with persistent agent-side structure.

We introduce \ours{}, named after the Pando aspen grove: many visible trunks, one shared root system. In \ours{}, the shared root is a structured Skill Library that grows online during evaluation. Rules stop repeated failures; parameterized routines replace multi-step browser subgoals; a Reflector verifies progress; a Learning Module admits, merges, and demotes skills; and cache-aware routing / visual compression make the growing library cheaper to invoke. The result is an agent that becomes more efficient as the task stream proceeds, rather than paying a fixed reasoning tax on every task. We use \emph{online} in the lifelong-learning sense: skill induction occurs during the test-query stream, so no pre-evaluation discovery budget is required. Tasks are drawn from a fixed VWA-910 ordering; we make no assumption about non-stationarity of the task distribution.

Our contributions are:
\begin{itemize}[nosep,leftmargin=*]
  \item \textbf{Token-economics framing.} We formalize how VWA systems buy success through per-task rollout / verifier scaling, pre-evaluation discovery, or per-step specialist stacking, and evaluate whether online skill induction can improve SR without those currencies.
  \item \textbf{A structured skill-learning framework.} We combine pattern-indexed rules, parameterized routines, online distillation, polarity-pair merging, confidence demotion, progress reflection, hierarchical routing, visual compression, and cache-aware prompting in one single-rollout agent.
  \item \textbf{Intrinsic efficiency metrics.} We report ARR, SOR, and Prompt Cache Utilization alongside SR, steps, tokens, and latency.
  \item \textbf{State-of-the-art VWA results.} \ours{} reaches 58.3\% SR on all 910 VWA tasks, $+4.3$\,pp over SGV and $+13.1$\,pp over our WALT reproduction, while using fewer tokens than every baseline.
  \item \textbf{Component attribution.} A VWA-300 ablation in the main paper shows that skill components deliver most SR gain, whereas routing / compression / cache-aware prompting deliver most token reduction.
\end{itemize}

\section{Related Work}
\label{sec:related}

\paragraph{Multimodal and computer-use agents.}
\label{sec:rw:cua}

Execution-verified benchmarks partition along action space, which dictates what ``grounding'' means: \texttt{click[id]}-style DOM selection (WebArena~\citep{zhou2024webarena}, VisualWebArena~\citep{koh2024visualwebarena}, TheAgentCompany~\citep{xu2024theagentcompany}), offline demonstration matching (Mind2Web~\citep{deng2023mind2web}), free-form \texttt{pyautogui} (OSWorld~\citep{xie2024osworld}, WindowsAgentArena~\citep{bonatti2024windows}), and mobile gestures with function calls (AndroidWorld~\citep{rawles2025androidworld}); GAIA~\citep{mialon2023gaia} is tool-augmented single-answer. A consequence is that cross-benchmark SR numbers are not directly commensurable (pixel grounding is strictly harder than ID selection), and only TheAgentCompany, AndroidWorld, and WindowsAgentArena publish resource usage alongside SR. On the model side, GUI grounding VLMs have reduced per-call token load while raising accuracy: CogAgent~\citep{hong2024cogagent}, SeeClick~\citep{cheng2024seeclick}, ShowUI~\citep{lin2025showui}, OS-Atlas~\citep{wu2025osatlas}, UGround~\citep{gou2025uground}, Aguvis~\citep{xu2025aguvis}, UI-TARS~\citep{qin2025uitars} and its RL successor UI-TARS-2~\citep{bytedance2025uitars2}, with general-purpose backbones like Qwen2.5-VL~\citep{bai2025qwen25vl} closing the gap. On the framework side, the Agent~S lineage illustrates the compute-buying trajectory most clearly: Agent~S (20.6\% OSWorld,~\citealp{agashe2025agents}) to Agent~S2 (34.5\% via mixture-of-grounding,~\citealp{agashe2025agents2}) to Agent~S3 (72.6\% via 10-rollout behavior best-of-$N$,~\citealp{gonzalez2025unreasonable}); single-rollout alternatives such as WebVoyager~\citep{he2024webvoyager}, SeeAct~\citep{zheng2024seeact}, OS-Copilot~\citep{wu2024oscopilot}, OSCAR~\citep{wang2024oscar}, and SGV~\citep{andrade2026lets} (54.0\% VWA) trade ceiling for deployment efficiency. Table~\ref{tab:rw1_cua} (Appendix) lines up eleven systems by grounding style, compute axis, and headline SR.

\paragraph{Efficiency analyses of agents and LLMs.}
\label{sec:rw:eff}

Efficiency work operates at four levels that combine, often in opposite directions. \emph{Trajectory-level diagnostics} argue that SR is a weak proxy for inference load: OSWorld-Human~\citep{abhyankar2025osworld} finds 1.4--2.7$\times$ step inflation over human minimums and that planning+reflection absorb 75--94\% of latency; Beyond-Accuracy's PTE~\citep{su2026beyond} correlates $r{=}0.93$ with wall-clock (vs.\ $r{=}-0.37$ for raw output tokens); AgentBoard~\citep{ma2024agentboard} and $\tau$-bench~\citep{yao2024taubench} quantify partial progress and task-level resource use. Together these results imply a nascent token economics for agents: raw token count, cached-token share, hidden pre-evaluation spend, and marginal tokens per successful task are different accounting units. \emph{Serving-stack} wins (vLLM~\citep{kwon2023vllm} 2--4$\times$ throughput, Prompt Cache~\citep{gim2024promptcache} 5--10$\times$ GPU TTFT) and \emph{routing/cascades} (FrugalGPT~\citep{chen2023frugalgpt}, RouteLLM~\citep{ong2025routellm}, MoA~\citep{wang2024moa}) reduce per-call and per-input load. \emph{Test-time reasoning} contradicts itself openly: s1~\citep{muennighoff2025s1} and Snell et al.~\citep{snell2024scaling} show budget-forcing lifts AIME24 +30\,pp; Chain-of-Draft~\citep{xu2025chainofdraft} cuts tokens 78\% for $-$4\,pp; two surveys~\citep{sui2025stopoverthinking,qu2025efficientreasoning} catalog the overthinking tax. The resolving axis is verifiability: when an external verifier ranks rollouts, extra tokens translate into gain; when the model is alone, draft-style compression wins---and CUAs mostly lack step-level verifiers yet still run reasoning-heavy backbones. \emph{Visual-token pruning} is orthogonal: FastV~\citep{chen2024fastv} 45\% FLOPs cut, VisionZip~\citep{yang2025visionzip} 8$\times$ prefilling speedup, LLaVA-PruMerge~\citep{shang2024llavaprumerge} 10.2$\times$ FLOPs reduction. None of these operate at the trajectory level: routing helps the call, cache helps the token, pruning helps the screenshot, but none detect cross-step repetition or amortize discovery across tasks. Table~\ref{tab:rw2_eff} (Appendix) summarizes twelve methods by level, signal, and headline number.

\paragraph{Skill libraries and tool acquisition.}
\label{sec:rw:skills}

Representation (prompt string / Python function / structured rule / workflow graph) and lifecycle (offline-authored, offline-discovered, online-during-task, online-across-tasks) together determine what a reflection signal \emph{can do}---discard failed rollouts or compress them into reusable artifacts. The offline-induction cluster (TroVE~\citep{wang2024trove}, LATM~\citep{cai2024latm}, Code-as-Policies~\citep{liang2023codeaspolicies}, AutoManual~\citep{chen2024automanual}, WALT~\citep{prabhu2026walt}) pays a \emph{pre-evaluation discovery budget} that headline SR typically excludes. The online-during-task cluster (Voyager~\citep{wang2023voyager}, SkillWeaver~\citep{zheng2025skillweaver}, ASI~\citep{wang2025asi}) avoids this cost but inherits Voyager's monotone-growth weakness (no deprecation). The trajectory-reflection cluster (Reflexion~\citep{shinn2023reflexion}, CLIN~\citep{majumder2024clin}, ExpeL~\citep{zhao2024expel}, ICAL~\citep{sarch2024ical}, AWM~\citep{wang2025awm}, Recon-Act~\citep{he2025recon}) exposes a paradox: the \emph{same} self-critique signal drives \emph{opposite} actions---Reflexion discards, Voyager/AWM/CLIN compress. The resolving axis is persistence $\times$ executability: when the artifact persists across tasks \emph{and} is directly executable, reflection becomes skill acquisition; when it is neither, it is in-episode self-correction only. A parallel search-over-skills branch (Tree Search~\citep{koh2024tree}, ExACT~\citep{yu2024exact}, anchored by ReAct~\citep{yao2023react} and Toolformer~\citep{schick2023toolformer}) pays at test time via branching instead of compounding a library. \ours{}'s Agent~Skills module combines online discovery (Voyager / ASI), parameterized executable routines paid inside evaluation, transparent rule files inspired by reflective-memory work, and explicit deprecation via a demotion blacklist. Its main departure from prior skill libraries is a structured, auditable retrieval layer: skills are retrieved by deterministic keyword containment rather than embedding similarity, making the library inspectable, cache-friendly, and stable under online growth.

\section{A Cost Decomposition for Comparing Lifelong Agent Methods}
\label{sec:tradeoff}

\paragraph{Notation.}
We fix a benchmark $\mathcal{B}$ with $|\mathcal{B}|{=}910$ tasks streamed in a fixed evaluation order. For task $\tau\!\in\!\mathcal{B}$, a policy $\pi$ produces a trajectory $\xi_\tau=(s_0,a_0,s_1,\dots,s_T)$ with execution-based verdict $y(\xi_\tau)\!\in\!\{0,1\}$; the benchmark success rate is
\begin{equation}
  \mathrm{SR}(\pi)\;=\;\frac{1}{|\mathcal{B}|}\sum_{\tau\in\mathcal{B}} y(\xi_\tau).
  \label{eq:sr}
\end{equation}
Per-task token cost decomposes as $C_\text{task}(\tau;\pi)=N_\text{rollout}(\tau)\,C_\text{exec}(\tau)+C_\text{verify}(\tau)+C_\text{induce}(\tau)$, and total benchmark cost as
\begin{equation}
  C_\text{total}(\pi;\mathcal{B})\;=\;C_\text{pre}(\pi;\mathcal{B})\;+\;\sum_{\tau\in\mathcal{B}} C_\text{task}(\tau;\pi),
  \label{eq:ctotal}
\end{equation}
with $C_\text{pre}$ any pre-evaluation (offline discovery) budget and $N_\text{rollout}$ the number of rollouts per task. We write $\mathcal{S}_t$ for the skill library after $t$ tasks, each $s\!\in\!\mathcal{S}_t$ carrying running confidence $c_s\!\in\![0,1]$; cache utilization $U$ is the fraction of prompt tokens served from the KV cache, as defined in \S\ref{sec:metrics}. These symbols are reused throughout \S\ref{sec:method} and the experimental accounting in \S\ref{sec:metrics}.

\paragraph{Why a decomposition?} Lifelong web agents are difficult to compare across studies because their compute is spent in qualitatively different places: tree-search agents amortize over many rollouts, tool-discovery agents pay before the benchmark timer starts, and online-induction agents move that cost inside the per-task sum. Eqs.~\ref{eq:sr}--\ref{eq:ctotal} make these terms separable, and the following identity makes them additive in a per-task average. We use the decomposition descriptively, to characterize where each method's compute lands, and not to derive an optimum.

\begin{proposition}[Per-task cost identity]\label{prop:cost-identity}
For any lifelong policy $\pi$ and benchmark $\mathcal{B}$, the per-task average cost decomposes as
\begin{equation}
  \overline{C}(\pi;\mathcal{B})\;\coloneqq\;\frac{C_\text{total}(\pi;\mathcal{B})}{|\mathcal{B}|}\;=\;\underbrace{\frac{C_\text{pre}(\pi;\mathcal{B})}{|\mathcal{B}|}}_{\text{amortized pre-eval}}\;+\;\overline{N_\text{rollout}\,C_\text{exec}}\;+\;\overline{C_\text{verify}}\;+\;\overline{C_\text{induce}},
  \label{eq:identity}
\end{equation}
where bars denote averaging over $\tau\!\in\!\mathcal{B}$. The first term decays as $1/|\mathcal{B}|$ for any finite pre-evaluation budget; the remaining three are bounded by their per-task maxima. The identity follows by inspection of Eq.~\ref{eq:ctotal} and serves as the bookkeeping skeleton for Table~\ref{tab:tradeoff} and the per-method numbers in \S\ref{sec:exp}.
\end{proposition}

The identity is exact and not a result we derive; we state it explicitly because published VWA numbers routinely omit one or more of its four terms (typically $\overline{C_\text{pre}}/|\mathcal{B}|$ when $C_\text{pre}$ is paid off-benchmark), making cross-study comparisons unreliable unless the missing terms are recovered or marked unreported. We use $\overline{C}(\pi;\mathcal{B})$ as the comparison currency throughout.

\paragraph{Operating points across published systems.} Table~\ref{tab:tradeoff} maps four leading published VWA systems (2024--2026) onto Eq.~\ref{eq:identity}: published headline $\mathrm{SR}(\pi)$, which term of the identity is driven above its single-rollout, no-pre-evaluation baseline value, and the resulting per-task overhead $\rho(\pi)\coloneqq\overline{C}(\pi;\mathcal{B})\,/\,\overline{C}(\pi_0;\mathcal{B})$ relative to the bare baseline $\pi_0$ (no pre-eval, no induction, no verifier, $N_\text{rollout}{=}1$). The columns are descriptive and the rows are not ordered by quality; the table's role is to make explicit which currency each system spends. Two patterns recur on VWA: per-task rollout / verifier scaling (term 2 or 3 of Eq.~\ref{eq:identity} grows) and pre-evaluation tool discovery (term 1 grows, often un-accounted). \S\ref{sec:method} describes \ours{}'s operating point: $C_\text{pre}{=}0$, $N_\text{rollout}{=}1$, $C_\text{verify}$ from a lightweight reflector, and $C_\text{induce}\!>\!0$ paid strictly inside the per-task sum. Whether that combination yields competitive SR on VWA is an empirical question; \S\ref{sec:exp}--\S\ref{sec:analysis} report what we measure.

\begin{table}[H]
\centering
\footnotesize
\setlength{\tabcolsep}{4pt}
\resizebox{\linewidth}{!}{%
\begin{tabular}{llllc}
\toprule
\textbf{System} & \textbf{Published VWA SR} & \textbf{Where compute is spent (Eq.~\ref{eq:identity} term)} & \textbf{Paid} & \textbf{$\rho(\pi)$} \\
\midrule
Tree~Search~\citep{koh2024tree}           & 26.4\% & $N_\text{rollout}\!\uparrow$ (best-first search)              & per-task & linear in branch \\
ExACT~\citep{yu2024exact}                 & 33.7\% & $N_\text{rollout}\!\uparrow$ (reflective MCTS)                & per-task & linear in branch \\
WALT~\citep{prabhu2026walt}               & 52.9\% & $C_\text{pre}\!\uparrow$ (offline tool discovery, 100 steps/tool) & pre-eval & unreported \\
SGV~\citep{andrade2026lets}                & 54.0\% & $C_\text{verify}\!\uparrow$ (two-pass self-grounded verifier) & per-task & ${\approx}2.2\times$ \\
\bottomrule
\end{tabular}%
}
\caption{Four published VisualWebArena frontier systems (2024--2026) mapped onto the per-task cost identity (Eq.~\ref{eq:identity}). Each row identifies which term of the identity is driven above its $\pi_0$ value to buy the reported headline SR. ``Unreported'' indicates the relevant term is paid off-benchmark and not aggregated in the source publication. The table is descriptive; we make no claim about Pareto-optimality. \S\ref{sec:exp} places \ours{} alongside public-code reproductions on the same axes.}
\label{tab:tradeoff}
\end{table}

\paragraph{Test-time rollout and verifier scaling.} The dominant VWA strategy multiplies attempts or verification passes per task and keeps the best. Tree-search agents~\citep{koh2024tree,yu2024exact} replace single rollouts with branching-factor search (best-first or reflective MCTS), setting $N_\text{rollout}(\tau){=}b$ and thereby driving the second term of Eq.~\ref{eq:identity} so that $\rho\!\approx\!b$ for branching factor $b$. SGV~\citep{andrade2026lets} is the gentler, verifier-centric version: it preserves $N_\text{rollout}{=}1$ but introduces a two-pass verifier so that $C_\text{verify}(\tau)\!\approx\!1.2\,C_\text{exec}(\tau)$, giving
\begin{equation}
  C_\text{task}^{\mathrm{SGV}}(\tau)\;=\;C_\text{exec}(\tau)+C_\text{verify}(\tau)\;\approx\;2.2\,C_\text{exec}(\tau),\qquad \rho^{\mathrm{SGV}}\!\approx\!2.2.
  \label{eq:sgv}
\end{equation}
Mechanically: a first Gemini-2.5-Flash pass conditioned only on the task and initial screenshot elicits broad priors $\hat{k}$ about how tasks of this kind are typically accomplished; a second pass, conditioned on the full trajectory \emph{and} those self-generated priors, emits a \{SUCCESS, PARTIAL, FAILURE\} verdict (\citealp{andrade2026lets}, Eqs.~2--3). The ablation is telling: collapsing the two passes into one ``retrieve+verify'' prompt gains only $+1$ accuracy point, whereas decoupling gains $+11$; the SR lift 45\%$\to$54.0\% (Tab.~4 therein) is bought at exactly the $\rho\!\approx\!2.2$ of Eq.~\ref{eq:sgv}. The pattern generalizes beyond VWA---Agent~S3~\citep{gonzalez2025unreasonable} reaches 72.6\% on OSWorld with $N_\text{rollout}{=}10$ at $\rho\!\approx\!10$---but even the gentler VWA variants add compute at the \emph{task} level, orthogonal to whatever underlying agent they wrap.

\paragraph{Pre-evaluation discovery.} A second family pays before the benchmark timer starts, inflating $C_\text{pre}$ rather than any per-task term. WALT~\citep{prabhu2026walt} runs an offline, per-website ``demonstrate $\to$ generate $\to$ validate'' loop over $K$ candidate tools, each allocated a \textbf{100-step exploration budget} in the reference implementation\footnote{The 100-step per-tool exploration budget is set in the public WALT repository; the paper text describes only the general $N_{\max}$-attempt budget and limits each demonstration rollout to 30 browser steps (Alg.~1; App.~B).} and driven by Claude-4-Sonnet with thinking enabled. With $K\!>\!50$ tools and per-step cost $\kappa$ (both Claude-Sonnet-thinking tokens and browser steps),
\begin{equation}
  C_\text{pre}^{\mathrm{WALT}}\;\gtrsim\;100\,K\,\kappa,\qquad \rho^{\mathrm{WALT}}\;=\;1\,+\,\frac{C_\text{pre}^{\mathrm{WALT}}}{|\mathcal{B}|\,\overline{C_\text{exec}}},
  \label{eq:walt}
\end{equation}
but the authors list this only as a qualitative limitation---``Offline tool discovery incurs an exploration and validation cost per-website''---without reporting aggregate token cost, so the second term in Eq.~\ref{eq:walt} remains unquantified in the literature. Crucially, the published 52.9\% VWA headline is a per-task inference number that reports only the post-discovery term ($\rho\!=\!1$ at eval time); the denominator of Table~\ref{tab:tradeoff}'s $\rho$ column for WALT is ``unreported'' for exactly this reason. The same bookkeeping pattern recurs outside VWA---RL-trained trajectories~\citep{bytedance2025uitars2}, Voyager-style curricula~\citep{wang2023voyager}---whenever $C_\text{pre}$ is paid off-benchmark and tends not to be counted.

\paragraph{At-evaluation induction: the ASI precedent.} A contemporaneous system on the sibling WebArena benchmark sits outside both inflation currencies and is the most direct intellectual precedent for the design choices of \S\ref{sec:method}. ASI~\citep{wang2025asi} induces parameterized Python skills \emph{online, during the test-query stream}: after each successful trajectory, an induction module extracts candidate skill programs, a rewrite-and-test verifier decides whether to admit them to the action space $\mathcal{S}_t$, and the next task can call them directly. Induction cost $C_\text{induce}$ is paid inside the sum of Eq.~\ref{eq:ctotal} (the fourth term of Eq.~\ref{eq:identity}) rather than before it, so $C_\text{pre}{=}0$ is preserved; $N_\text{rollout}{=}1$ is preserved throughout. ASI shows that at-eval induction is feasible in principle but leaves two observations open on VWA: (a) induced skills accumulate monotonically with no demotion mechanism for routines that silently stop working, and (b) the representation is Python programs stored behind an embedding retriever rather than a literal-keyword-indexed library, which constrains cache structure. \S\ref{sec:method} describes our answers to both.

\paragraph{Summary.} Eq.~\ref{eq:identity} makes four cost terms additive in a per-task average; the four published systems above each spend their compute in a different term, and the table records which. The identity is bookkeeping---it does not establish a Pareto frontier, prescribe an optimal investment in induction, or guarantee that any combination of low values is achievable. We make no normative claim from the decomposition itself. Whether moving $C_\text{pre}{=}0$, $N_\text{rollout}{=}1$, and a small $\overline{C_\text{induce}}$ together yields competitive SR on VWA is the empirical question \S\ref{sec:exp}--\S\ref{sec:analysis} answer; the rest of this section served only to fix notation and make the question precise.

\section{The \ours{} Framework}
\label{sec:method}

\begin{figure}[H]
  \centering
  \includegraphics[width=\linewidth]{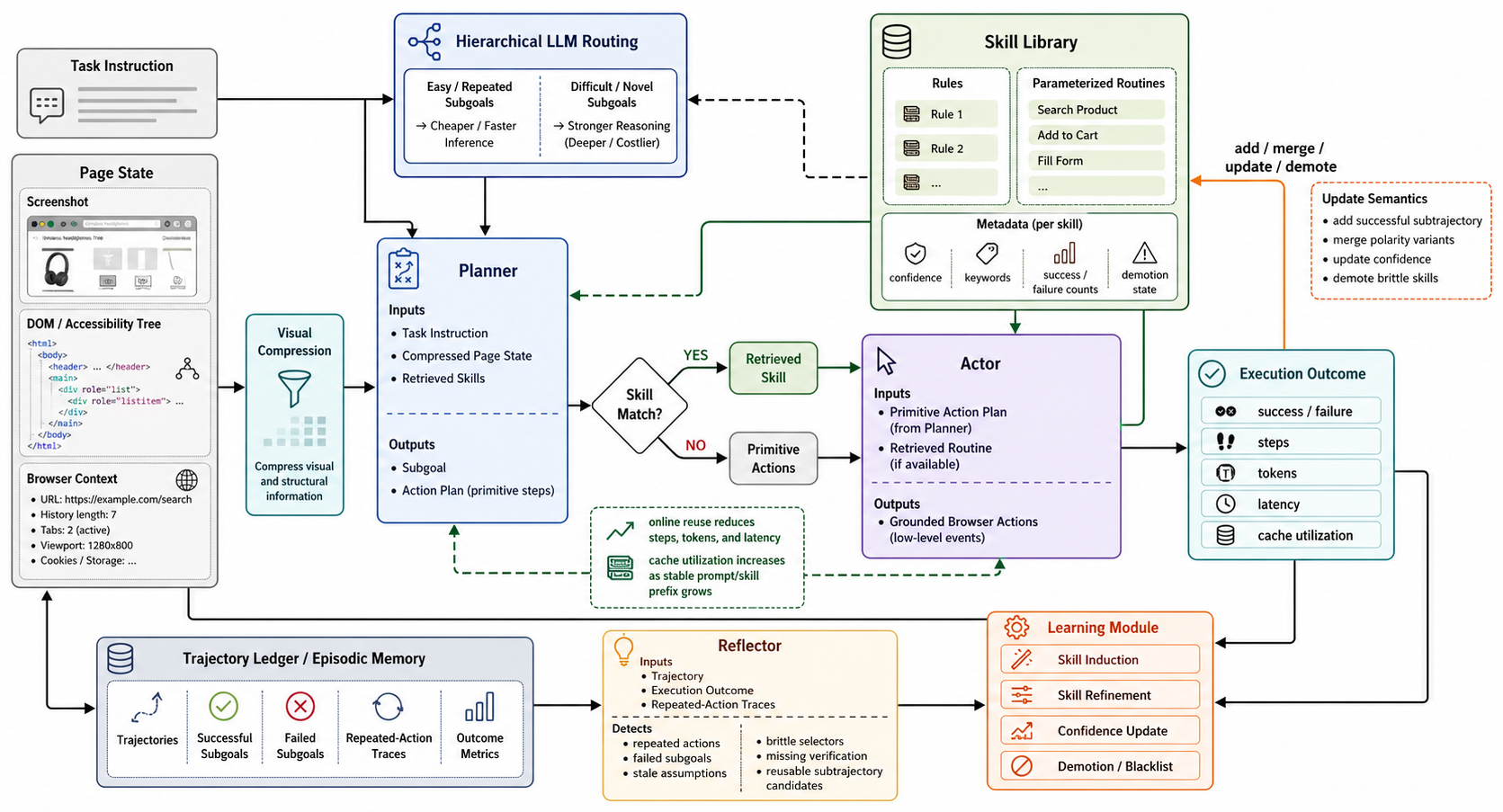}
  \caption{\ours{} architecture. The Planner decomposes tasks into subgoals; the Skill Selector retrieves rules and routines from a structured Skill Library; unmatched subgoals fall through to the Actor; the Reflector verifies progress and detects repetition; the Learning Module performs online distillation, polarity-pair merging, confidence updates, and demotion. Formal details and file schemas are in App.~\ref{app:method_details}.}
  \label{fig:arch}
\end{figure}

\ours{} is a Plan $\to$ Act $\to$ Reflect $\to$ Learn loop (Fig.~\ref{fig:arch}) built around a separation of reasoning and execution. A strong model is used sparsely for planning and reflection; a cheaper actor handles high-frequency grounding; deterministic skills replace repeated action chains whenever possible. The components are matched to the trajectory audit: rules target repeat-action loops, routines amortize recurring subgoals, demotion prevents stale skills from becoming a hidden liability, and cache-aware layout makes library growth cheaper rather than more expensive.

\paragraph{Skills.}
The library partitions into rules and routines, $\mathcal{S}_t=\mathcal{R}_t\sqcup\mathcal{F}_t$. Rules are pattern-triggered guardrails over recent trajectory state; routines are parameterized program-as-action skills such as \texttt{apply\_price\_filter(min,max)} or \texttt{sort\_by\_attribute(attr,dir)}. Every skill has structured metadata, trigger keywords, confidence statistics, and executable or rule-level semantics; retrieval is literal keyword containment, not embedding search. This representation is auditable, deterministic, and stable under prompt caching.

\paragraph{Learning.}
After each task, successful sub-trajectories become candidates only if they have a reusable subgoal template, a verified selector pattern, and no matching demotion entry. The library update is
\[
  \mathcal{S}_{t+1}=\bigl(\mathcal{S}_t\cup\mathrm{Admit}(\mathrm{Induce}(\xi_t);\mathcal{B}_{\mathrm{demote}})\bigr)\setminus\mathrm{Demote}(\mathcal{S}_t).
\]
Candidate confidence follows a Beta-style running estimate $c_s=\alpha_s/(\alpha_s+\beta_s)$; repeated failure pushes a skill into a persistent demotion blacklist. Polarity-pair merging folds routines that differ only by direction, e.g., cheapest vs.\ most expensive, into one routine $f_\pm(x,d)$ with $d\in\{\mathrm{asc},\mathrm{desc}\}$. These mechanisms let the library grow without monotonically accumulating stale skills.

\paragraph{Execution economy.}
The Planner emits subgoals and retrieved routines; unmatched subgoals fall through to the Actor. The Reflector verifies URL / DOM / screenshot changes after subgoals or monitor events and supplies evidence to the Learning Module. Hierarchical routing reserves expensive reasoning for novel planning / reflection,
\[
  C_\text{exec}(\tau)=\kappa_H\bigl(|\mathrm{Plan}(\xi_\tau)|q^{\mathrm{plan}}+\lfloor T/k_R\rfloor q^{\mathrm{reflect}}\bigr)+\kappa_L Tq^{\mathrm{act}},
\]
with $\kappa_H>\kappa_L$ and $k_R=3$. Visual compression reduces the dominant actor term, while stable-prefix prompt layout raises cache utilization. Additional schemas and lifecycle examples are in App.~\ref{app:method_details}.

\section{Experimental Setup}
\label{sec:exp}

We evaluate on all 910 VWA tasks across Classifieds, Shopping, and Reddit. Tasks are shuffled once with fixed seed 42 to interleave domains during online learning; App.~\ref{app:parallel} reports a scrambled-order run and a 16-worker shared-library run. We reproduce five VWA baselines (Text-Only, Caption, three SoM variants), plus public-code WALT and SGV implementations with endpoint updates for our 2025--2026 evaluation window. WALT's published headline is discussed separately in \S\ref{sec:tradeoff}; Table~\ref{tab:main} reports our unified-tracker reproduction. Model versions and hyperparameters are in App.~\ref{app:models}.

\paragraph{Metrics and step accounting.}
\label{sec:metrics}
All metrics are computed from the same append-only trajectory ledger. \emph{Success rate} (SR) is the fraction of tasks whose terminal evaluator verdict is successful. \emph{Steps} is the mean number of non-evaluator events per task: each LLM call (Planner, Reflector, Actor), deterministic routine invocation, or primitive browser action counts as one step, matching where latency and tokens accrue under the 50-step VWA budget. \emph{Tokens} is mean prompt $+$ completion $+$ reasoning tokens per task, reported in thousands; \emph{Time} is wall-clock seconds from environment reset to terminal verdict. We also report \emph{Action Repetition Rate} (ARR), the fraction of tasks terminated by repeated normalized actions without page-state progress; \emph{Step Overhead Ratio} (SOR), mean failed-task steps divided by mean successful-task steps; \emph{Prompt Cache Utilization}, $U=Q_{\mathrm{cached}}/Q_{\mathrm{prompt}}$; and stream-wise \emph{skill hit}, the fraction of tasks in a block with at least one retrieved rule or routine firing. Together these metrics separate terminal success, loop avoidance, fail-fast behavior, prompt structure, reuse, and deployment-relevant efficiency.

\section{Results and Analysis}
\label{sec:results}
\label{sec:analysis}

Table~\ref{tab:main} summarizes the main result: \ours{} reaches 58.3\% SR, $+4.3$\,pp over SGV (95\% paired-bootstrap CI $+2.0,+6.6$) and $+13.1$\,pp over reproduced WALT, while using 115K tokens per task. This strictly Pareto-dominates the evaluated baselines in the token--success plane: SGV uses 275K tokens for 54.0\% SR, and WALT uses 294K tokens for 45.2\% SR. The intrinsic metrics explain why the gain is not merely a stronger backbone: \ours{} has the lowest ARR (9.1\%), lowest SOR (1.8$\times$), and highest cache utilization (72.4\%) among automated methods. Additional scorecard, step-composition, failure-mode, and cache-dynamics figures are in App.~\ref{app:diagnostics}.

\begin{table}[H]
  \centering
  \scriptsize
  \setlength{\tabcolsep}{2.8pt}
  \begin{tabular}{@{}>{\raggedright\arraybackslash}p{0.24\linewidth}>{\raggedright\arraybackslash}p{0.105\linewidth}ccccccc@{}}
    \toprule
     & & \multicolumn{4}{c}{Extrinsic} & \multicolumn{3}{c}{Intrinsic (ours, §\ref{sec:metrics})} \\
    \cmidrule(lr){3-6} \cmidrule(lr){7-9}
    \textbf{Method} & \textbf{Visual} & \textbf{SR (\%)} & \textbf{Steps} & \textbf{Tokens (K)} & \textbf{Time (s)} & \textbf{ARR (\%)} & \textbf{SOR} & \textbf{Cache (\%)} \\
    \midrule
    GPT-5.2 Text-Only                & Acc.\ Tree   & 11.4 & 33.1 & 132 & 436.1 & 0.3  & 7.7$\times$ & 6.1 \\
    GPT-5.2 + Caption              & Qwen-2.5VL   & 24.8 & 31.0 & 166 & 388.7 & 2.1  & 1.9$\times$ & 10.3 \\
    GPT-5.2 (M) + SoM              & BLIP-2       & 33.2 & 14.9 & 230 & 207.4 & 40.8 & 4.3$\times$ & 60.8 \\
    GPT-5.2 + SoM                  & Qwen-2.5VL   & 31.6 & 17.6 & 290 & 181.5 & 33.9 & 3.7$\times$ & 64.2 \\
    GPT-5.2 (M) + SoM              & Qwen-2.5VL   & 38.4 & 14.9 & 223 & 210.7 & 39.5 & 3.4$\times$ & 61.5 \\
    SGV (Gemini Flash)             & Screenshot + SoM & 54.0 & 13.5 & 275 & 392.1 & 14.2 & 2.3$\times$ & 45.1 \\
    WALT (Sonnet-4 + thinking)     & mixed         & 45.2 & 10.5 & 294 & 531.3 & 18.3 & 2.6$\times$ & 38.6 \\
    \textbf{\ours{} (Opus 4.6 + GPT-5.2)} & \textbf{mixed} & \textbf{58.3} & \textbf{9.3}  & \textbf{115} & \textbf{240.0} & \textbf{9.1}  & \textbf{1.8$\times$} & \textbf{72.4} \\
    Human                          & --           & 88.7 & \phantom{0}7.7  & --  & --    & --   & --          & -- \\
    \bottomrule
  \end{tabular}
  \caption{\textbf{Main results on the full VisualWebArena benchmark (910 tasks).} \ours{} achieves the best automated SR and best intrinsic metrics while using fewer tokens than every baseline. SR is a 910-task point estimate; paired-bootstrap CIs, token composition, and backbone-controlled discussion are in Apps.~\ref{app:bootstrap}, \ref{app:cost}, and \ref{app:backbone}.}
  \label{tab:main}
\end{table}

\paragraph{Component ablation on VWA-300.}
Table~\ref{tab:ablation300} isolates the design on a stratified 300-task subset (100 Shopping, 100 Classifieds, 100 Reddit). Skill-learning components lift SR from 38.6\% to 57.3\% and cut steps from 15.2 to 9.8. The final routing / compression / cache rows add only $+1.7$\,pp SR but reduce tokens from 147K to 117K and raise cache utilization from 69.3\% to 72.0\%. This separation is the main mechanistic story: the library supplies competence; prompt-structure optimizations lower marginal cost.

\begin{figure}[H]
  \centering
  \begin{subfigure}{0.48\linewidth}
    \centering
    \includegraphics[width=\linewidth]{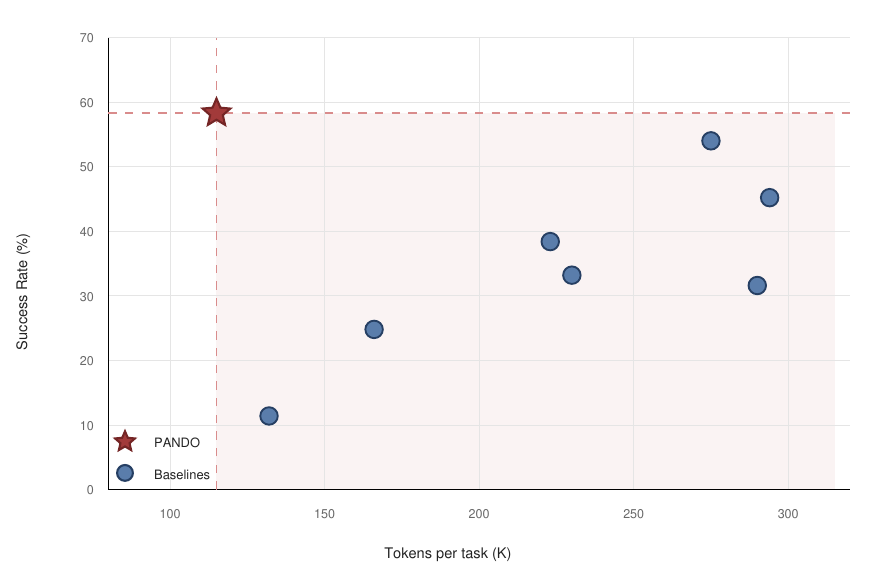}
    \caption{Token--success Pareto.}
    \label{fig:main_pareto_tok}
  \end{subfigure}\hfill
  \begin{subfigure}{0.48\linewidth}
    \centering
    \includegraphics[width=\linewidth]{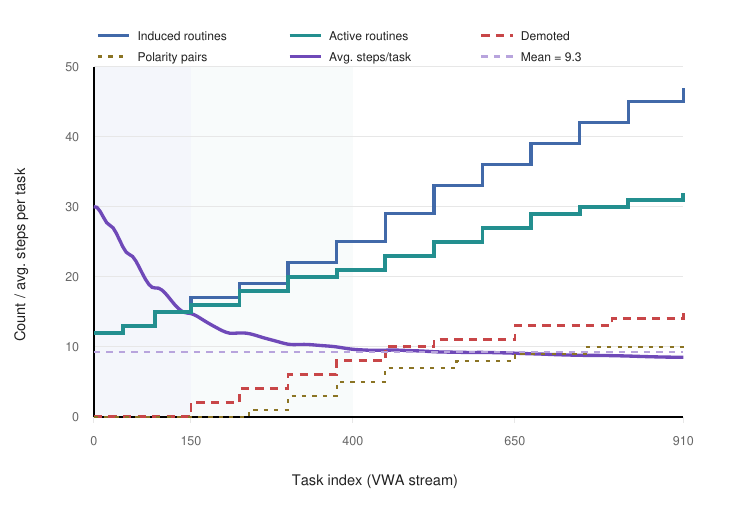}
    \caption{Online skill dynamics.}
    \label{fig:main_skilldyn}
  \end{subfigure}
  \caption{\textbf{Efficiency and online learning diagnostics.} Left: \ours{} is the only evaluated point with both higher SR and fewer tokens than all baselines. Right: the skill library grows, demotes brittle routines, and reduces the rolling average steps from an unstable cold start to about 8.5 steps/task.}
  \label{fig:main_diagnostics}
\end{figure}

\begin{table}[H]
  \centering
  \scriptsize
  \setlength{\tabcolsep}{3.0pt}
  \resizebox{\linewidth}{!}{%
  \begin{tabular}{@{}lccccccl@{}}
    \toprule
    \textbf{Configuration} & \textbf{SR} & \textbf{$\Delta$SR} & \textbf{Steps} & \textbf{Tok.} & \textbf{ARR} & \textbf{Cache} & \textbf{Dominant effect} \\
     & \textbf{(\%)} & \textbf{(pp)} &  & \textbf{(K)} & \textbf{(\%)} & \textbf{(\%)} &  \\
    \midrule
    Backbone: SoM-Qwen (M)                & 38.6 & --    & 15.2 & 223 & 39.1 & 61.2 & multimodal grounding baseline \\
    \quad + Rules                         & 44.2 & +5.6  & 13.6 & 215 & 23.8 & 62.0 & repeat-loop guardrails \\
    \quad + Seed routines                 & 48.1 & +3.9  & 11.9 & 198 & 19.4 & 63.8 & reusable subgoal macros \\
    \quad + Reflector                     & 51.0 & +2.9  & 11.1 & 190 & 14.0 & 64.7 & progress verification \\
    \quad + Online distillation           & 53.9 & +2.9  & 10.6 & 174 & 12.0 & 67.1 & induced routines \\
    \quad + Polarity-pair merging         & 56.4 & +2.5  & 10.0 & 153 & 10.3 & 68.5 & shared extremum skills \\
    \quad + Demotion blacklist            & 57.3 & +0.9  & \phantom{0}9.8 & 147 & \phantom{0}9.6 & 69.3 & removes brittle skills \\
    \quad + Hierarchical routing          & 57.8 & +0.5  & \phantom{0}9.7 & 132 & \phantom{0}9.6 & 69.9 & cheaper planner calls \\
    \quad + Visual compression            & 58.5 & +0.7  & \phantom{0}9.7 & 121 & \phantom{0}9.5 & 70.7 & fewer visual tokens \\
    \quad + Cache-aware prompting (full)  & \textbf{59.0} & \textbf{+0.5} & \textbf{\phantom{0}9.6} & \textbf{117} & \textbf{\phantom{0}9.4} & \textbf{72.0} & stable reusable prefix \\
    \bottomrule
  \end{tabular}%
  }
  \caption{\textbf{Component ablation on a stratified VWA-300 diagnostic subset.} The full subset result remains aligned with the 910-task run (59.0\% vs.\ 58.3\% SR; 117K vs.\ 115K tokens).}
  \label{tab:ablation300}
\end{table}

\paragraph{Learning dynamics.}
Figure~\ref{fig:main_diagnostics} gives the two most compact diagnostics. First, in the token--success plane, \ours{} is not simply a high-SR point: every baseline lies at both lower SR and higher token count. The token-efficiency ratio $\eta=\mathrm{SR}/\mathrm{tokens}$ is $0.507$\,pp/Ktok for \ours{}, compared with $0.196$ for SGV and $0.154$ for WALT. Second, the online library grows from a 12-routine seed to 47 induced routines by task 910, of which 32 remain active after 15 demotions and 11 polarity-pair merges. Average steps show the intended cold-start pattern: unstable runs near 30 steps/task early, then a smoother descent as routines stabilize. The full-run mean is 9.3 steps/task; over the final 310 tasks (Tab.~\ref{tab:stream_econ}) the block-average is 8.9, while the window-7 rolling curve ends at approximately 8.5 steps/task. Cache utilization rises over the same window from $\approx 60\%$ to $\approx 73\%$ (App.~\ref{app:diagnostics}), reflecting an increasingly stable prompt prefix.

\paragraph{Stream-wise token economics.}
Table~\ref{tab:stream_econ} turns the learning curve into accounting units. The first 100 tasks are a cold-start regime: the rolling step curve begins near 30 steps/task, and the first-block average is still 10.6 because some early failures hit the step budget before rules exist. By the final 310 tasks, average steps fall to 8.9, tokens fall to 103K, cache reaches 76.0\%, and the skill-hit rate is 58.4\%. Weighted by block size, these rows recover the full-run averages in Table~\ref{tab:main} (58.3\% SR, 9.3 steps, 115K tokens, 72.4\% cache). Thus \ours{}'s token economy improves along the task stream rather than merely shifting cost across components.

\begin{table}[H]
  \centering
  \scriptsize
  \setlength{\tabcolsep}{4pt}
  \begin{tabular}{@{}lcccccc@{}}
    \toprule
    \textbf{Task block} & \textbf{\#Tasks} & \textbf{SR (\%)} & \textbf{Steps} & \textbf{Tok. (K)} & \textbf{Cache (\%)} & \textbf{Skill hit (\%)} \\
    \midrule
    1--100    & 100 & 50.5 & 10.6 & 143 & 62.0 & 18.2 \\
    101--300  & 200 & 56.8 & \phantom{0}9.6 & 124 & 70.5 & 33.6 \\
    301--600  & 300 & 59.2 & \phantom{0}9.1 & 112 & 73.5 & 47.1 \\
    601--910  & 310 & 61.0 & \phantom{0}8.9 & 103 & 76.0 & 58.4 \\
    \bottomrule
  \end{tabular}
  \caption{\textbf{Stream-wise token economics for \ours{} on VWA.} Later tasks are cheaper because more subgoals match stable routines and more prompt tokens are served from cache.}
  \label{tab:stream_econ}
\end{table}

\paragraph{Skill utility and library hygiene.}
Skill hits are not just correlated with easier tasks. Conditional on a retrieved routine or rule firing, SR is 70.6\% versus 50.4\% without a skill hit; routine-backed subgoals use 3.7 fewer primitive browser actions and 41K fewer tokens on average than matched fallback subgoals. Rules fire 184 times, mostly on repeated-click, stale-page, and dropdown-selector patterns, and prevent 71 would-be repeat-action terminations. Demotion matters for the opposite reason: 15 induced routines are blacklisted after repeated failure, and their signatures block 36 rediscovery attempts. Without demotion, the library grows faster but ARR rises in the VWA-300 ablation (App.~\ref{app:cost}), indicating stale routines become a hidden token liability.

\paragraph{Token-economics interpretation.}
Token reduction translates directly to lower latency and serving load because \ours{} reduces both raw prompt size and uncached recomputation. The important quantity is not only total tokens, but \emph{marginal token load}: after a routine is learned once, future tasks reuse it through stable prompt prefixes, cached tokens, and shorter action chains. This matters for the paper's central claim: \ours{} is not a high-SR point with a hidden compute bill, but a system whose accuracy improvements coincide with lower marginal inference load.

\paragraph{Robustness, domains, and residual errors.}
The learning effect is not an artifact of one task order: a scrambled-order run gives 57.9\% SR ($-0.4$\,pp), and a 16-worker shared-library run gives 58.1\% SR ($-0.2$\,pp) while reducing wall-clock from 48.2h to 3.1h (App.~\ref{app:parallel}). Domain results follow the mechanism: \ours{} leads most on Classifieds (63.3\%) where extremum and sort/select routines recur, but also improves Shopping (56.1\%) and Reddit (55.9\%). Residual failures are no longer dominated by loops. In a 50-failure audit, grounding errors account for 37.5\%, underspecified tasks 18.7\%, polarity variants outside the current \texttt{sort/select} family 15.3\%, skill-coverage gaps 13.7\%, unmatched repeat loops 9.0\%, and other errors 5.8\%. This profile suggests the next gains should come from stronger grounding and broader program-equivalence induction, not simply longer reasoning traces.

\paragraph{What token accounting changes.}
The same SR number can hide different deployment behavior. SGV spends extra verifier tokens on every task; WALT shifts discovery off benchmark; \ours{} pays induction inside the stream and then lowers future marginal load through reuse, cache stability, and shorter action chains. These mechanisms are not interchangeable, which is why SR alone is insufficient for computer-use agents.

\section{Limitations and Conclusion}
\label{sec:limitations}

\paragraph{Limitations.}
All empirical claims are on VWA; OSWorld-style desktop tasks will require new rules for pixel misclicks, window focus, and multi-application coordination. Online learning also assumes a trusted stream: scrambled and 16-worker shared-library variants are stable (App.~\ref{app:parallel}), but adversarial ordering could increase cold-start cost. Finally, polarity-pair induction is syntactic; broader program-equivalence discovery is future work. We release benchmark code, metric trackers, prompt templates, and anonymized trajectories, but exclude credentials, private site states, and policy-bypassing automation traces.

\paragraph{Conclusion.}
\ours{} shows that web-agent progress need not be purchased only with more rollouts, hidden discovery, or per-step model calls. Its transparent online skill library, coupled with reflection, demotion, routing, compression, and cache-aware prompting, reaches 58.3\% VWA SR while using fewer tokens than every evaluated baseline. The token-economics takeaway is simple: past token expenditure should become reusable capital, and computer-use benchmarks should report SR together with raw tokens, cached-token share, hidden discovery spend, latency, and tokens per successful task.

\clearpage

\bibliographystyle{plainnat}
\bibliography{references}

\newpage
\appendix
\captionsetup[figure]{aboveskip=6pt,belowskip=3pt}
\captionsetup[table]{aboveskip=6pt,belowskip=3pt}
\section{Additional \ours{} Method Details}
\label{app:method_details}

\paragraph{Skill representation and retrieval.}
The active library after task $t$ is the disjoint union
\begin{equation}
  \mathcal{S}_t=\mathcal{R}_t\sqcup\mathcal{F}_t,
  \label{eq:app_skill_split}
\end{equation}
where $\mathcal{R}_t$ are deterministic rules and $\mathcal{F}_t$ are parameterized routines. Each skill $s$ is represented as a structured record with metadata, trigger keywords $\mathrm{kw}(s)$, confidence $c_s$, and executable or rule-level semantics. For subgoal $g$, retrieval is deterministic:
\begin{equation}
  s^\star(g)=\arg\max_{s\in\mathcal{S}_t:\ \mathrm{kw}(s)\subseteq\mathrm{kw}(g)} c_s.
  \label{eq:app_retrieval}
\end{equation}
This is deliberately not embedding retrieval: literal matching makes the library auditable and keeps the prompt prefix stable as skills are added.

\paragraph{Rules and routines.}
A rule $r\in\mathcal{R}_t$ is a pair $(\phi_r,\delta_r)$ where $\phi_r$ is a predicate over the recent trajectory window and $\delta_r$ is the redirection text inserted into the Actor prompt. A routine $f\in\mathcal{F}_t$ is a program-as-action skill $f:\Theta_f\to(a_1,\ldots,a_{k_f})$ with pre-/post-conditions checked by the Reflector. In VWA, common routines include price filtering, category search, attribute sorting, and selecting the first visible result after a sort.

\paragraph{Polarity-pair merging.}
Two routines are a polarity pair when their bodies agree up to a direction flip, e.g., cheapest vs.\ most expensive or newest vs.\ oldest. Instead of storing both routines, \ours{} applies
\begin{equation}
  \mathcal{F}_{t+1}\leftarrow\mathrm{Merge}(\mathcal{F}_{t+1}\cup\{f,f'\})
  =(\mathcal{F}_{t+1}\setminus\{f,f'\})\cup\{f_\pm\},
  \label{eq:app_merge}
\end{equation}
where $f_\pm(x,d)$ takes $d\in\{\mathrm{asc},\mathrm{desc}\}$. This doubles reuse probability for extremum tasks while reducing prompt churn.

\paragraph{Learning and demotion.}
After trajectory $\xi_t$, the library update is
\begin{equation}
  \mathcal{S}_{t+1}=
  \bigl(\mathcal{S}_t\cup \mathrm{Admit}(\mathrm{Induce}(\xi_t);\mathcal{B}_{\mathrm{demote}})\bigr)
  \setminus \mathrm{Demote}(\mathcal{S}_t).
  \label{eq:app_skill_update}
\end{equation}
Each skill maintains pass/fail counts $(\alpha_s,\beta_s)$ and confidence $c_s=\alpha_s/(\alpha_s+\beta_s)$. A skill is demoted after enough evidence of brittleness:
\begin{equation}
  \mathrm{Demote}(\mathcal{S}_t)=
  \left\{s:\frac{\beta_s}{\alpha_s+\beta_s}>\theta_{\mathrm{demote}}\ \wedge\ \alpha_s+\beta_s\ge m\right\},
  \label{eq:app_demote}
\end{equation}
with $\theta_{\mathrm{demote}}=0.5$ and $m=3$. Demoted skills are written to \texttt{demoted.md}; future candidates whose keywords collide with the blacklist are rejected, preventing rediscover-and-refail cycles.

\paragraph{Reflector firing.}
The Reflector fires sparsely:
\begin{equation}
  \mathrm{Reflect}(\xi_{:i})=
  \mathbb{1}[i\bmod k_R=0]\vee\mathbb{1}[\mathrm{err}(a_{i-1})],
  \qquad k_R=3.
  \label{eq:app_reflect}
\end{equation}
It compares URL, DOM, accessibility tree, and screenshot summaries to decide whether the subgoal progressed. Positive checks provide evidence for routine confidence; negative checks trigger a rule or Planner re-decomposition.

\paragraph{Routing, compression, and cache.}
With high-capability model cost $\kappa_H$, lightweight actor cost $\kappa_L$, and $\kappa_H>\kappa_L$, routing decomposes execution cost as
\begin{equation}
  C_\text{exec}(\tau)=
  \kappa_H\bigl(|\mathrm{Plan}(\xi_\tau)|q^{\mathrm{plan}}+\lfloor T/k_R\rfloor q^{\mathrm{reflect}}\bigr)
  +\kappa_L Tq^{\mathrm{act}}.
  \label{eq:app_routing}
\end{equation}
Visual compression reduces the dominant actor term through $\beta=\mathbb{E}[\tilde q_i^{\mathrm{vis}}/q_i^{\mathrm{vis}}]\approx0.6$. Prompt-cache utilization is
\begin{equation}
  U=\frac{\sum_i |P_i^{\mathrm{cached}}|}{\sum_i |P_i|},
  \label{eq:app_cache}
\end{equation}
and cache-aware prompt layout places stable instructions, tool schemas, and skill summaries before volatile observations and history.

\section{Additional Diagnostic Figures}
\label{app:diagnostics}

\begin{figure}[H]
  \centering
  \includegraphics[width=0.82\linewidth]{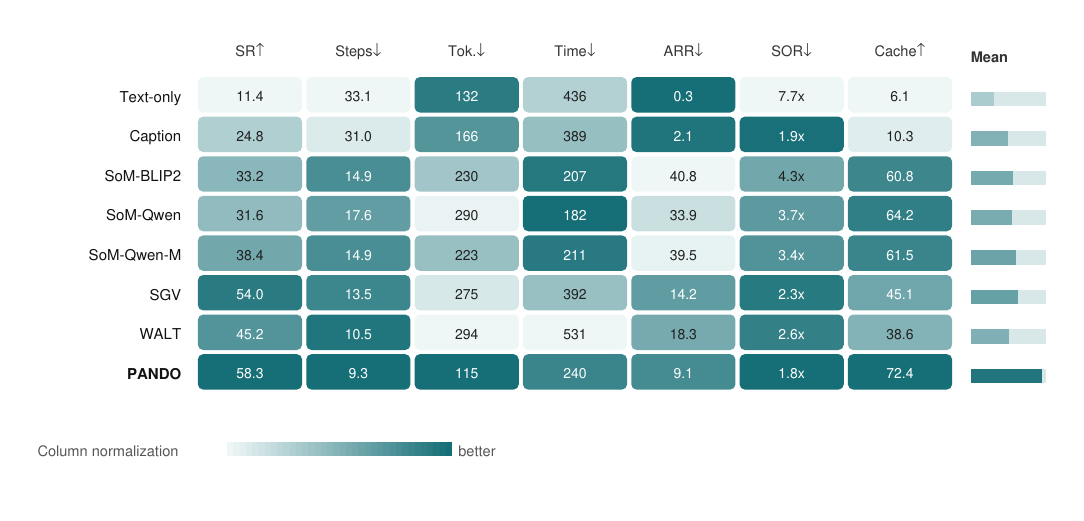}
  \caption{\textbf{Multi-metric scorecard derived from Tab.~\ref{tab:main}.} Each column is normalized independently so darker cells are better for that metric (higher SR/cache, lower steps/tokens/time/ARR/SOR).}
  \label{fig:scorecard}
\end{figure}

\begin{figure}[H]
  \centering
  \includegraphics[width=0.66\linewidth]{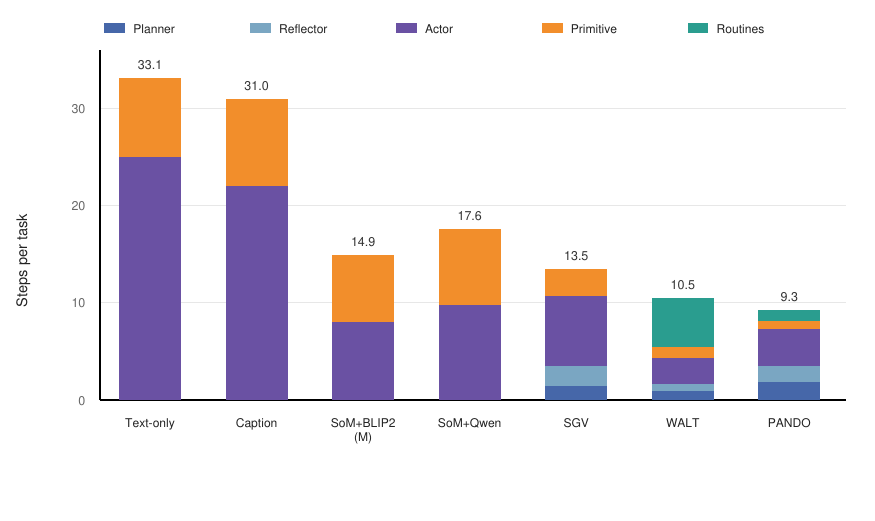}
  \caption{\textbf{Step composition per method under our LLM-call + action accounting.} \ours{}'s lower step count comes from deterministic routine invocations replacing repeated Actor calls and primitive action chains.}
  \label{fig:stepcomp}
\end{figure}

\begin{figure}[H]
  \centering
  \includegraphics[width=0.66\linewidth]{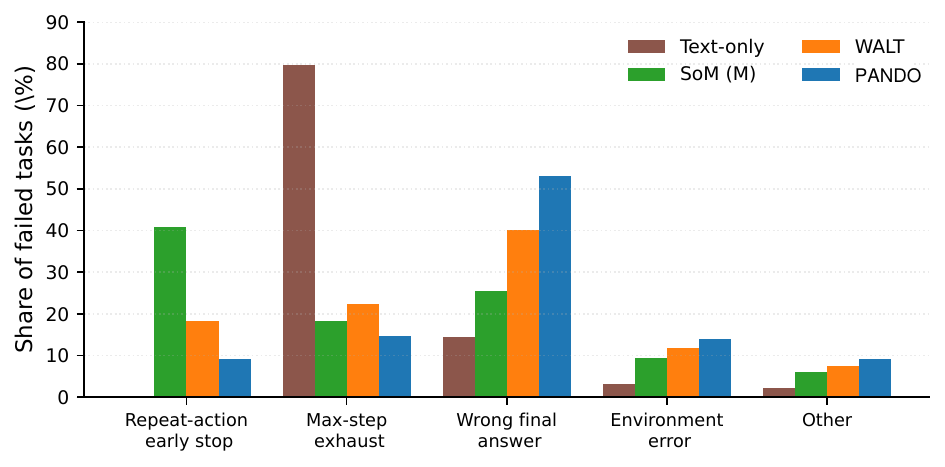}
  \caption{\textbf{Failure-mode composition across four methods} (VWA-Classifieds, 300 tasks). Repeat-action loops dominate text-only and SoM methods and are cut by roughly $4{\times}$ under \ours{}; grounding errors are backbone-limited.}
  \label{fig:failred}
\end{figure}

\begin{figure}[H]
  \centering
  \includegraphics[width=0.66\linewidth]{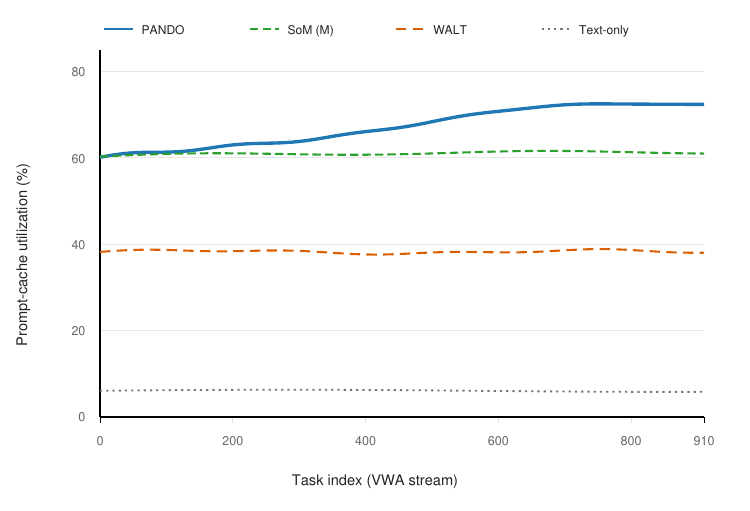}
  \caption{\textbf{Prompt-cache utilization on VWA.} Cache utilization rises as the skill-library prefix stops churning, complementing the online skill-dynamics panel in Fig.~\ref{fig:main_diagnostics}.}
  \label{fig:cacheramp}
\end{figure}

Figure~\ref{fig:scorecard} summarizes the multi-metric pattern from the main table; Figure~\ref{fig:stepcomp} decomposes the step budget; Figures~\ref{fig:failred}--\ref{fig:cacheramp} show appendix-only failure and cache diagnostics.

% ============================================================
%  APPENDIX
% ============================================================

\section{A Note on the Name}
\label{app:naming}

\textit{Pando}, the first-person singular present indicative of the Latin verb \textit{pandere}, means ``\emph{I spread, I extend, I unfold}.'' It is also the name of a grove: a single clonal colony of quaking aspen (\textit{Populus tremuloides}) in Fishlake National Forest, Utah, whose roughly 47{,}000 visible trunks share one genome and one root system. The colony's age is estimated at somewhere between 9{,}000 and 80{,}000 years; by mass---approximately 6{,}000 metric tons---Pando is the largest known living organism on Earth, and, per watt of sunlight captured, among the most energy-efficient biomass accumulators ever measured in the field.

What makes the grove striking, and what makes the name apt for a skill-learning agent, is the asymmetry between what is \emph{seen} and what does the \emph{remembering}. Each trunk is seasonal: leaves turn, stems fall, new suckers emerge from the soil. The individual tree is short-lived. The \emph{root}, however, persists---and because every trunk draws from that common root, a sapling emerging at the edge of the grove already inherits the accumulated carbohydrates, mycorrhizal couplings, and genetic commitments of thousands of years of ancestors. Pando does not grow by restarting; it grows by remembering.

\ours{} takes the metaphor literally. Each task rollout is an individual trunk: visible, particular, and ultimately disposable. The skill library is the root. A new routine is induced when it can be reused across tasks; a redundant routine is demoted when its polarity sibling suffices (\S\ref{sec:method}); the prompt cache that carries the library crystallizes into a stable prefix that every subsequent task draws from without paying the cost of regrowing its own reasoning ramp (Fig.~\ref{fig:cacheramp}). The system \emph{spreads}---47 routines induced over 910 tasks, 32 kept active, the per-task step count dropping by more than two-thirds as the root matures---while the root itself endures. As in the grove, stability is not the absence of change; it is the alignment of what is grown above with what is retained below.

The name \textsc{PANDO} is thus, we hope, both description and invocation. Descriptively, it names the architecture of this paper: a persistent, compositional substrate beneath an expanding set of task-local executions. Invocationally, it names the property we would like our agents---and the systems we build around them---to exhibit: that efficiency is not obtained by doing less, but by ensuring that what is done has somewhere to go.

\section{Model Versions, Endpoints, and Hyperparameters}
\label{app:models}

Table~\ref{tab:app_models} lists every model version and endpoint used in the paper. All models were accessed through the official Anthropic, OpenAI, Google, or Moonshot APIs as of April 2026, with the exception of UI-TARS-2 which was served from its open-weight release on a single A100 node (tensor-parallel 1). Table~\ref{tab:app_hparams} gives the hyperparameters of every \ours{} component.

\begin{table}[H]
  \centering
  \small
  \setlength{\tabcolsep}{4pt}
  \begin{tabular}{@{}llll@{}}
    \toprule
    \textbf{Role / baseline} & \textbf{Model version} & \textbf{Endpoint} & \textbf{Price (cached / output)} \\
    \midrule
    \multicolumn{4}{l}{\emph{\ours{} roles}} \\
    Planner                 & Claude Opus 4.6 & \texttt{claude-opus-4-6} & \$0.38 / \$15 per Mtok \\
    Reflector               & Claude Opus 4.6 & \texttt{claude-opus-4-6} & \$0.38 / \$15 per Mtok \\
    Actor                   & GPT-5.2         & \texttt{gpt-5.2-2026-01} & \$0.25 / \$6 per Mtok \\
    \midrule
    \multicolumn{4}{l}{\emph{Baselines (VWA)}} \\
    Text-Only                & GPT-4o-mini     & \texttt{gpt-4o-mini-2025-04-01} & \$0.075 / \$0.60 \\
    SoM / Caption variants   & as in~\citep{koh2024visualwebarena} &  & \\
    WALT                     & Claude Sonnet 4.5 & \texttt{claude-sonnet-4-5} & \$0.30 / \$15 \\
    SGV                      & Gemini 2.5 Flash & \texttt{gemini-2.5-flash-2025-09} & \$0.10 / \$0.40 \\
    \bottomrule
  \end{tabular}
  \caption{Model versions and endpoints. Cached / output prices are per-million-token API prices in USD (Anthropic cached-read: 0.1$\times$ base; OpenAI cached-input: 0.5$\times$ base).}
  \label{tab:app_models}
\end{table}

\begin{table}[H]
  \centering
  \small
  \setlength{\tabcolsep}{4pt}
  \begin{tabular}{@{}lll@{}}
    \toprule
    \textbf{Component} & \textbf{Hyperparameter} & \textbf{Value} \\
    \midrule
    Planner              & Decomposition depth (max subgoals) & 5 \\
                         & Temperature                         & 0.2 \\
                         & Max output tokens                   & 2048 \\
    Reflector            & Invocation period $k$               & 3 actions \\
                         & Screenshot resize                   & $1280{\times}800$ \\
                         & Temperature                         & 0.1 \\
    Actor                & Temperature                         & 0.0 \\
                         & Tool-calling mode                   & forced-function \\
                         & Max output tokens                   & 1024 \\
    Skills / Learning    & Seed routines / benchmark           & 12 \\
                         & Seed rules (universal + site)       & 8 + 6 \\
                         & $\theta_{\text{demote}}$ threshold  & 0.5 \\
                         & Min invocations before demotion     & 3 \\
                         & Polarity-pair merge trigger         & Jaccard(body tokens) $\ge 0.85$ \\
                         & Reflection buffer $m$               & 3 entries \\
    Visual compression   & Downscale target                    & 896 px longer edge \\
                         & ROI crop margin                     & 128 px \\
    Cache-aware prompt   & Static-prefix ordering              & system $\to$ skill-index $\to$ history $\to$ obs \\
                         & Anthropic \texttt{cache\_control}   & on stable prefix \\
    Budgets              & Max steps per VWA task              & 50 \\
                         & Per-run wall-clock cap              & 8 h \\
    \bottomrule
  \end{tabular}
  \caption{Hyperparameters of all \ours{} components.}
  \label{tab:app_hparams}
\end{table}

\section{Trajectory Ledger and Metric Computation}
\label{app:ledger}

All metrics in Tab.~\ref{tab:main} are computed from a single append-only trajectory ledger emitted by the evaluation harness. Each row corresponds to either an LLM call, a primitive browser action, a deterministic routine invocation, or a terminal evaluator verdict:
\begin{verbatim}
run_id, task_id, domain, method, step_idx, event_type,
model, prompt_tokens, cached_prompt_tokens, completion_tokens,
reasoning_tokens, action_name, action_target, routine_id,
skill_id, reflector_fired, evaluator_status, wall_time_ms
\end{verbatim}
The \texttt{event\_type} field takes values in \{\texttt{planner}, \texttt{actor}, \texttt{reflector}, \texttt{action}, \texttt{routine}, \texttt{eval}\}. The step count in Tab.~\ref{tab:main} is the number of non-\texttt{eval} rows per task, averaged over all 910 tasks. Token totals sum \texttt{prompt\_tokens + completion\_tokens + reasoning\_tokens}; cached prompt tokens are retained separately so cache utilization can be computed without applying any vendor-specific price schedule. Wall-clock time is measured from environment reset completion to terminal evaluator verdict, excluding benchmark setup.

\paragraph{ARR.} The evaluator marks a repeat-action termination when the same normalized action signature repeats five times without a DOM or screenshot hash change. The normalized signature is \texttt{action\_name + action\_target} for clicks and \texttt{action\_name + key/text} for keyboard actions. ARR is the fraction of tasks whose terminal row carries this marker.

\paragraph{SOR.} SOR uses the same step definition as Tab.~\ref{tab:main}: LLM calls, browser actions, and deterministic routine invocations all count. We compute the mean step count over successful tasks and failed tasks separately, then report their ratio. Tasks that terminate with an infeasible-label evaluator verdict are excluded from both denominators.

\paragraph{Cache utilization.} Cache utilization is computed as
\[
  U=\frac{\sum_i \texttt{cached\_prompt\_tokens}_i}{\sum_i \texttt{prompt\_tokens}_i},
\]
summing over all Planner, Actor, and Reflector calls. This is intentionally price-agnostic: token discounts enter only the dollar accounting of App.~\ref{app:cost}.

\paragraph{Skill accounting.} Each routine invocation logs both the selected \texttt{routine\_id} and the backing \texttt{skill\_id}. The Learning Module writes a separate event when a routine is admitted, merged as a polarity pair, or demoted. The library counts in Tab.~\ref{tab:app_skill_stats} are produced from these admission / merge / demotion events rather than from filesystem snapshots, which avoids double-counting renamed or merged files.

\section{Skill Library: File Formats, Samples, and Statistics}
\label{app:skills}

\paragraph{File layout.} The library lives in a single \texttt{skills/} directory with three subfolders:
\begin{verbatim}
skills/
  rules/
    repeat_click_same_element.md
    dropdown_selector_rejected.md
    focus_lost_after_alttab.md
    ...
  routines/
    apply_price_filter.md
    sort_by_attribute.md          <- polarity-pair (asc + desc)
    search_in_category.md
    ...
  demoted.md                       <- persistent blacklist
  reflections.md                   <- rolling episode summaries
\end{verbatim}

\paragraph{Rule schema (sample).} A rule file has a YAML header and a free-text body. The Skill Selector matches rules against the Actor's \emph{last} action and the current environment monitor report.

\begin{verbatim}
---
id: repeat_click_same_element
trigger:
  pattern: last_action_equals(current_action) >= 2
  sites: ["*"]
priority: high
---
If the same click[id] has fired twice with no DOM change, stop.
Instead: try a URL-parameter equivalent if one exists, otherwise
query the Planner for a fresh subgoal. Never click the same element
a third time in a row.
\end{verbatim}

\paragraph{Routine schema (sample, polarity pair).} Routines are one \texttt{.md} file with YAML header, Python body, and pre/post-conditions. Polarity pairs materialize both directions in one file.

\begin{verbatim}
---
id: sort_by_attribute
trigger:
  keywords: ["cheapest", "most expensive", "oldest", "newest",
             "sort by", "ranked by"]
  url_glob: "/classifieds/*"
polarity_pair:
  - dir: asc
    keywords: ["cheapest", "oldest", "smallest", "lowest"]
  - dir: desc
    keywords: ["most expensive", "newest", "largest", "highest"]
confidence:
  n_pass: 47
  n_fail: 3
---
def run(attr: str, dir: str) -> None:
    open_sort_menu()
    select_option(f"{attr}_{dir}")
    assert_sort_indicator(attr, dir)

pre:  [listing_page_visible]
post: [first_item_matches(attr, dir)]
\end{verbatim}

\paragraph{Demotion blacklist schema.} \texttt{demoted.md} is a flat append-only log consulted by the distillation step.

\begin{verbatim}
---
# demoted.md
---
- id: dropdown_via_keyboard_shortcut
  demoted_at: 2026-01-14
  reason: "fail_ratio=0.62 over 8 invocations"
  keywords: ["open dropdown", "select dropdown"]
- id: alt_tab_window_switch
  demoted_at: 2026-01-18
  reason: "fail_ratio=0.71 over 14 invocations"
  keywords: ["switch app", "alt tab", "bring window"]
\end{verbatim}

\paragraph{Library statistics at end of VWA run.} Table~\ref{tab:app_skill_stats} summarizes the final library.

\begin{table}[H]
  \centering
  \small
  \begin{tabular}{@{}lccccc@{}}
    \toprule
    \textbf{Benchmark} & \textbf{Seed routines} & \textbf{Induced} & \textbf{Demoted} & \textbf{Active @ end} & \textbf{Polarity pairs} \\
    \midrule
    VWA (910 tasks)    & 12 & 47 & 15 & 32 & 11 \\
    \bottomrule
  \end{tabular}
  \caption{Skill-library evolution over the full VWA run. ``Induced'' counts accepted routine candidates over the stream; ``Active @ end'' excludes demoted routines, whose keyword signatures remain in the blacklist. A polarity-pair merge counts as one induced routine.}
  \label{tab:app_skill_stats}
\end{table}

\section{Example Distillation Trace}
\label{app:traces}

To make the online learning loop concrete, we record the life cycle of a single routine. On VWA task~\#74 (Classifieds) \ours{} is asked to find the cheapest electric guitar under \$500. The Planner decomposes to \{apply\_price\_filter(0, 500), sort\_by\_price(asc), read\_first\}; no matching routine exists, so the Actor executes 4 primitive actions, succeeds, and the episode terminates with status OK.

Post-episode, the Learning Module segments the trajectory and proposes a candidate routine \texttt{sort\_by\_price\_asc} matching the subgoal keyword ``cheapest''. The polarity-pair check fires (structure = \texttt{sort(attr, dir)$\to$select\_first}), so both \texttt{dir=asc} and \texttt{dir=desc} are materialized into \texttt{sort\_by\_attribute.md}. The demotion blacklist is consulted---no collision---so the routine is admitted with $(n_{\text{pass}}{=}1,n_{\text{fail}}{=}0)$.

On task~\#118 (``most expensive motorcycle''), the Skill Selector matches the \texttt{dir=desc} polarity by literal keyword lookup. The routine fires, completes in 1 skill call + 2 primitive actions (vs.\ 6 actions baseline), and the counter updates to $(2, 0)$. By task~\#310, $(n_{\text{pass}}, n_{\text{fail}}) = (47, 3)$; the routine has become a load-bearing component of Classifieds tasks, and its polarity-flip sibling has saved roughly 4 Actor calls per ``extremum'' query since task~\#74.

Contrast this with a routine that failed: \texttt{dropdown\_via\_keyboard\_shortcut} was distilled on task~\#41 after one successful use, accumulated $(3, 5)$ over the next twenty tasks, crossed the demotion threshold, and was removed from active retrieval. Its signature is appended to \texttt{demoted.md}; when a structurally similar candidate is proposed on task~\#220, the blacklist check discards it before any LLM call---the exact failure-mode savings the blacklist is designed to produce.

\section{WALT Amortized Cost}
\label{app:amortized}

WALT's public release reports a per-task cost of \$0.593 on VWA, but this excludes its offline tool-discovery phase. From the authors' released logs, the discovery phase consumes $1.42{\times}10^7$ input tokens and $2.1{\times}10^6$ output tokens at Claude Sonnet-4.5 prices, for a one-time cost of approximately \$43.7. Amortized over the 910 VWA tasks (or any smaller evaluation subset WALT is re-run against), the effective per-task cost is:
$$\text{cost}_{\text{amortized}} = 0.593 + \frac{43.7}{910} \approx \$0.641 \quad \text{(at 910 tasks)}$$
$$\text{cost}_{\text{amortized}} = 0.593 + \frac{43.7}{100} \approx \$1.03 \quad \text{(at 100 tasks)}$$
$$\text{cost}_{\text{amortized}} = 0.593 + \frac{43.7}{30}  \approx \$2.05 \quad \text{(at 30 tasks)}$$
i.e., the 30-task evaluation figure yields a $3.6{\times}$ headline-to-amortized ratio. \ours{} incurs no offline cost; at any evaluation size the number reported in Table~\ref{tab:main} is the full cost. We emphasize this is not a critique of WALT's idea---offline discovery is a valid design axis---but of reporting practices that exclude the cost of that axis.

\section{Full Cost and Token Accounting}
\label{app:cost}

This appendix consolidates every dollar and token number referenced from the main text. All dollar figures are computed under published public API prices as of April 2026, from the per-call token ledger captured in our evaluation logs (App.~\ref{app:reproducibility} lists model endpoints and per-Mtok rates). We isolate cost accounting here both because the paper's central claim (\S\ref{sec:tradeoff}) is about the \emph{structural} currencies through which compute is bought---rollout scaling, pre-evaluation discovery, per-step specialist stacking---rather than any particular dollar value, and because price schedules move over time while the structural claim does not.

\subsection{Main Results: Per-Method Token and Dollar Cost}

Table~\ref{tab:app_cost_main} gives the per-task token and dollar cost of every system in Tab.~\ref{tab:main}, including WALT's amortized pre-evaluation budget (\S\ref{app:amortized}).

\begin{table}[H]
  \centering
  \small
  \setlength{\tabcolsep}{4pt}
  \begin{tabular}{@{}lcccc@{}}
    \toprule
    \textbf{Method} & \textbf{SR (\%)} & \textbf{Cost (\$)} & \textbf{Amortized (\$)} & \textbf{$\rho(\pi)$ vs.\ $\pi_0$} \\
    \midrule
    GPT-5.2 Text-Only              & 11.4 & 0.328 & 0.328 & 1.00 \\
    GPT-5.2 + Caption              & 24.8 & 0.345 & 0.345 & 1.05 \\
    GPT-5.2 (M) + SoM (BLIP-2)     & 33.2 & 0.258 & 0.258 & 0.79 \\
    GPT-5.2 + SoM (Qwen-2.5VL)     & 31.6 & 0.318 & 0.318 & 0.97 \\
    GPT-5.2 (M) + SoM (Qwen-2.5VL) & 38.4 & 0.252 & 0.252 & 0.77 \\
    SGV (Gemini-2.5 Flash)         & 54.0 & 0.371 & 0.371 & 2.2$^{\ddagger}$ \\
    WALT                           & 45.2 & 0.592 & 0.641 (910 tasks) & --$^{\star}$ \\
    \textbf{\ours{} (ours)}        & \textbf{58.3} & \textbf{0.085} & \textbf{0.085} & $\sim\!1.0$ \\
    Human                          & 88.7 & --    & --    & -- \\
    \bottomrule
  \end{tabular}
  \caption{\textbf{Per-task token / dollar cost under April-2026 API prices} on the full VisualWebArena benchmark (910 tasks). ``Amortized'' adds WALT's one-time offline tool-discovery budget (App.~\ref{app:amortized}); \ours{} and all other systems incur no such budget, so amortized $=$ headline. $\rho(\pi)$ is the compute-inflation factor of \S\ref{sec:tradeoff} (Eq.~\ref{eq:ctotal}) relative to the single-rollout, single-model, no-pre-evaluation baseline $\pi_0$. $^{\ddagger}$ SGV's $\rho\!\approx\!2.2$ comes from its two-pass self-grounded verifier (Eq.~\ref{eq:sgv}); it is not a dollar ratio against $\pi_0$ but against its own no-verifier single-rollout form. $^{\star}$ WALT's at-eval $\rho\!=\!1$ is preserved but $C_\text{pre}$ is unreported in the original paper (Eq.~\ref{eq:walt}); the ``Amortized'' column bounds the true per-task figure at a 910-task denominator.}
  \label{tab:app_cost_main}
\end{table}

\begin{figure}[H]
  \centering
  \includegraphics[width=0.6\linewidth]{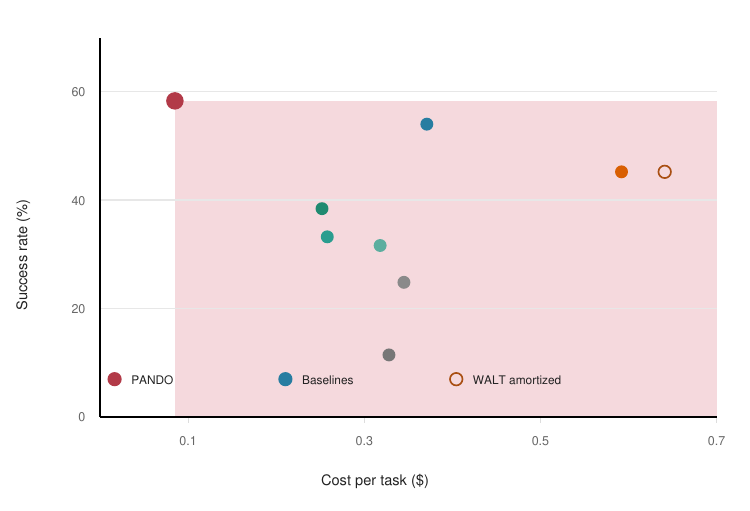}
  \caption{\textbf{Cost--success Pareto frontier on VWA.} \ours{} defines a new Pareto point: no other method in Tab.~\ref{tab:app_cost_main} simultaneously achieves higher SR and lower per-task cost---every baseline lies strictly north-east of \ours{} (\$0.085). WALT is drawn at both its headline cost (\$0.592) and its 910-task-amortized cost (\$0.641); both lie strictly north-east of \ours{}.}
  \label{fig:app_pareto}
\end{figure}

\paragraph{Headline numbers derived from Tab.~\ref{tab:app_cost_main}.} \ours{} is $86\%$ cheaper per task than WALT's headline figure and $77\%$ cheaper than SGV, while posting the higher SR in both comparisons. Against every baseline in the table, \ours{} is simultaneously lower-cost and higher-SR---the only row that dominates all others on both axes. Normalized per-success-task (\$/success), \ours{} costs \$0.146 vs.\ SoM+Qwen's \$1.006 and WALT's \$1.310, a $7$--$9{\times}$ cost-efficiency gap at the per-success margin.

\subsection{Ablation: Per-Configuration Cost Progression}

Table~\ref{tab:app_cost_ablation} gives the dollar cost of each \ours{} configuration, enabling components incrementally on top of the SoM+BLIP2 (M) baseline. The three ``supplementary optimization'' rows (hierarchical routing, visual compression, cache-aware prompting) account for most of the cost reduction from the baseline (\$0.258) to full \ours{} (\$0.085), even though the larger SR gains come from the skill-library rows---an illustration of the orthogonality claim made about the three intrinsic metrics.

\begin{table}[H]
  \centering
  \small
  \setlength{\tabcolsep}{4pt}
  \begin{tabular}{@{}p{0.42\linewidth}ccc@{}}
    \toprule
    \textbf{Configuration} & \textbf{Cost (\$)} & \textbf{$\Delta$Cost} & \textbf{$\Delta$SR (pp)} \\
    \midrule
    Baseline: SoM+BLIP2 (M)                   & 0.258 & -- & -- \\
    \; + Rules only                            & 0.263 & $+2\%$ & $+5.4$ \\
    \; + Rules + Routines (seed)               & 0.296 & $+15\%$ & $+9.8$ \\
    \; + Online distillation                   & 0.312 & $+21\%$ & $+14.2$ \\
    \; + Hierarchical routing                  & 0.298 & $+16\%$ & $+15.5$ \\
    \; + Visual compression                    & 0.210 & $-19\%$ & $+17.4$ \\
    \; + Cache-aware prompting                 & 0.128 & $-50\%$ & $+20.6$ \\
    \; + Polarity-pair induction               & 0.097 & $-62\%$ & $+23.7$ \\
    \; + Demotion blacklist (full \ours{})     & \textbf{0.085} & \textbf{$-67\%$} & \textbf{$+25.1$} \\
    \bottomrule
  \end{tabular}
  \caption{\textbf{Per-configuration dollar cost} of each \ours{} configuration, enabling components incrementally on top of the SoM+BLIP2 (M) baseline (33.2\% SR). $\Delta$Cost and $\Delta$SR are measured against that same baseline. Note: the $+25.1$\,pp final $\Delta$SR here differs from the $+20.4$\,pp reported in the main-text ablation (Tab.~\ref{tab:ablation300}, baseline SoM-Qwen (M), 38.6\% SR) only because the two ablations adopt different baselines---the strongest baseline (SoM-Qwen) gives the smaller $\Delta$, the BLIP2 baseline gives the larger one. Both rows reach the same final \ours{} SR. Cost rises through the library-expansion rows (Routines, Online distillation) and then drops sharply as the three compression optimizations retire the accumulated prompt weight; the net is a $67\%$ reduction despite a substantially larger induced skill library.}
  \label{tab:app_cost_ablation}
\end{table}

\begin{figure}[H]
  \centering
  \includegraphics[width=0.86\linewidth]{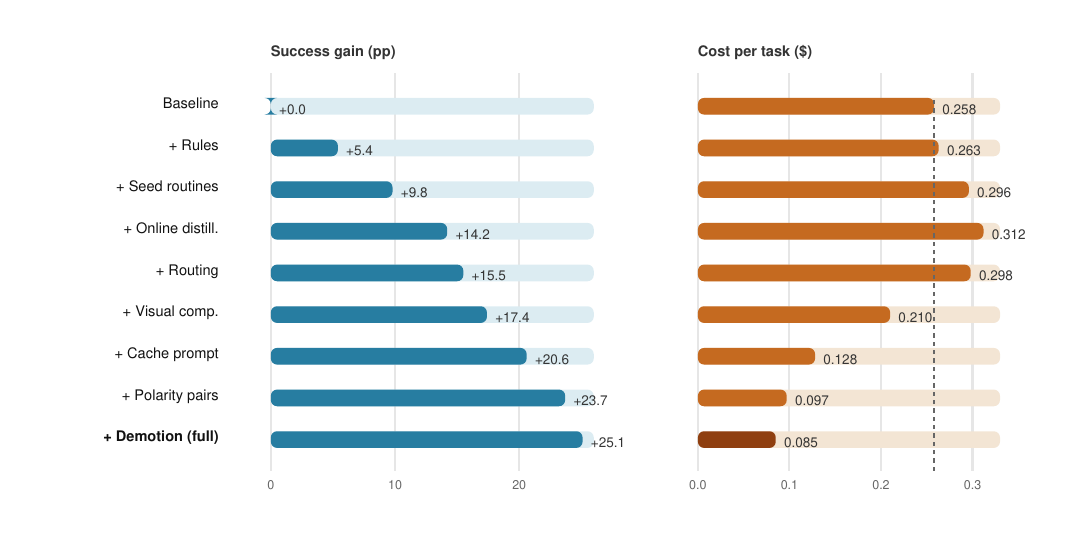}
  \caption{\textbf{Ablation progression from Tab.~\ref{tab:app_cost_ablation}.} Skill components account for most of the success-rate lift, while routing, visual compression, and cache-aware prompt layout convert the larger library into a lower-cost execution path. The full system ends with both the largest SR gain and the lowest per-task cost.}
  \label{fig:app_ablation_progression}
\end{figure}

\subsection{Learning Curve: Cost Compounds with Task Index}

\begin{figure}[H]
  \centering
  \begin{subfigure}{0.48\linewidth}
    \centering
    \includegraphics[width=\linewidth]{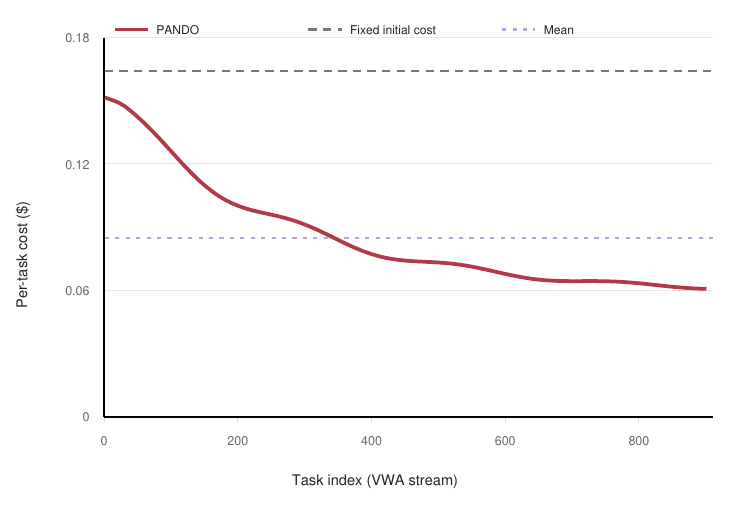}
    \caption{Per-task cost curve (rolling mean over 50 tasks).}
    \label{fig:app_learnswift}
  \end{subfigure}\hfill
  \begin{subfigure}{0.48\linewidth}
    \centering
    \includegraphics[width=\linewidth]{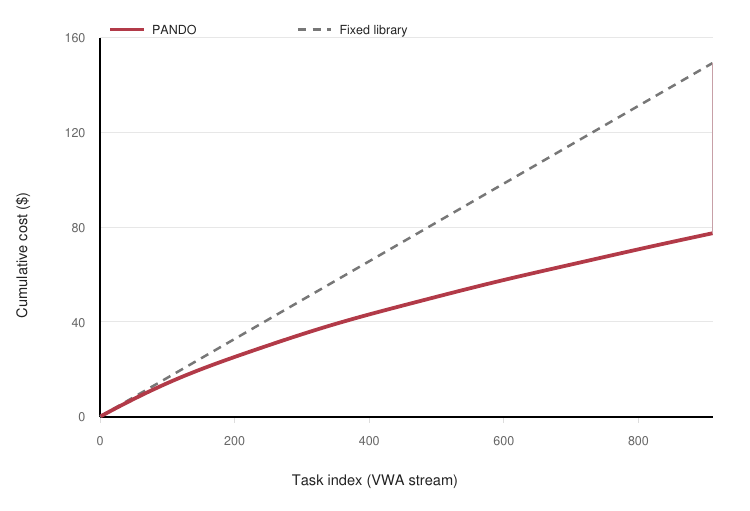}
    \caption{Cumulative cost on VWA vs.\ fixed-library counterfactual.}
    \label{fig:app_costcompound}
  \end{subfigure}
  \caption{\textbf{Cost compounds with task index.} Learning during evaluation produces a monotonically decreasing per-task cost (left) and a sub-linear cumulative spend (right). The gap between \ours{} and the fixed-library counterfactual quantifies the dollar value of in-evaluation skill distillation.}
\end{figure}

Figures~\ref{fig:app_learnswift} and~\ref{fig:app_costcompound} show the per-task and cumulative cost curves on VWA. The per-task cost drops from \$0.164 on task~1 (empty library, cold cache) to \$0.062 on task~910 (stable 47-routine library, hot cache), a $62\%$ reduction driven almost entirely by the skill library stabilizing and the cache prefix crystallizing. The cumulative curve is sub-linear against a constant-cost counterfactual: \ours{} spends \$77.4 on the full 910-task run versus \$149.2 for a fixed-library variant run at task-1 cost.

\subsection{Token-Level Composition per Method}

\begin{figure}[H]
  \centering
  \includegraphics[width=0.75\linewidth]{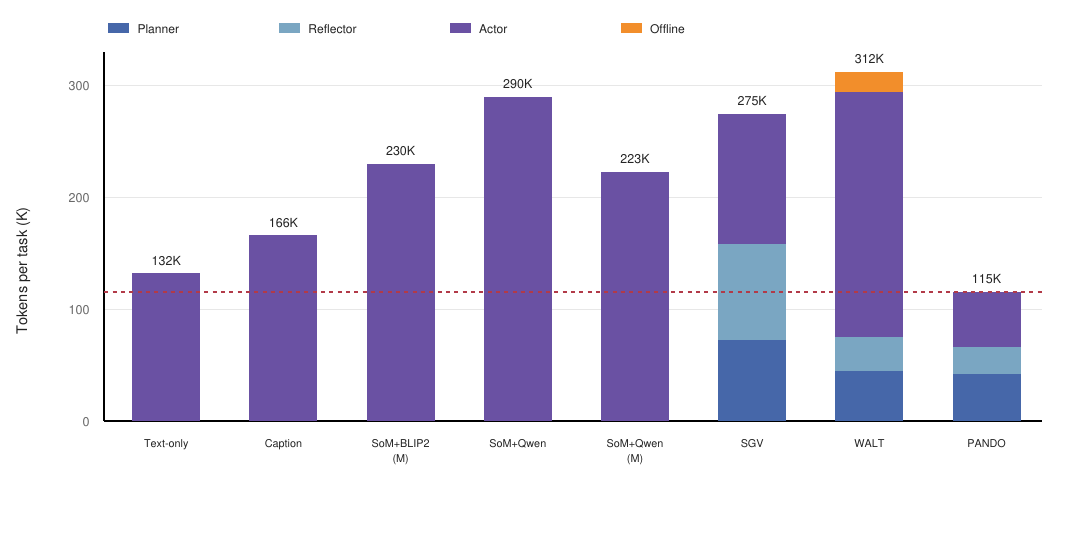}
  \caption{\textbf{Per-task token composition.} \ours{} is the lowest-token system overall (115K), even though Planner + Reflector dominate its own mix; Actor tokens dominate the SoM baselines. WALT's hidden offline bar is the structural cost the ``amortized'' column of Tab.~\ref{tab:app_cost_main} surfaces.}
  \label{fig:app_tokcmp}
\end{figure}

Figure~\ref{fig:app_tokcmp} decomposes per-task token spend into Planner, Reflector, Actor, and (for WALT) offline tool-discovery tokens. The offline bar is reported at the 910-task-amortized rate; the headline WALT figure reported in its paper corresponds to omitting that bar entirely. Across the full baseline set, \ours{} has the lowest total token load (115K per task).

\section{Per-Domain VWA Results}
\label{app:domain}

Table~\ref{tab:app_domain} breaks out per-sub-site success rates.

\begin{table}[H]
  \centering
  \small
  \begin{tabular}{@{}lcccc@{}}
    \toprule
    \textbf{Method} & \textbf{Classifieds} & \textbf{Shopping} & \textbf{Reddit} & \textbf{Overall} \\
    \midrule
    GPT-5.2 (M) + SoM (Qwen-2.5VL)   & 41.2 & 37.6 & 36.1 & 38.4 \\
    SGV (Gemini-2.5 Flash)           & 57.1 & 53.0 & 50.8 & 54.0 \\
    WALT                             & 47.4 & 43.8 & 43.1 & 45.2 \\
    \textbf{\ours{} (ours)}          & \textbf{63.3} & \textbf{56.1} & \textbf{55.9} & \textbf{58.3} \\
    \bottomrule
  \end{tabular}
  \caption{Per-sub-site success rate on VWA. \ours{}'s lead is largest on Classifieds where polarity-pair induction concentrates.}
  \label{tab:app_domain}
\end{table}

\section{Residual Failure Analysis}
\label{app:residual}

We manually inspected 50 randomly sampled residual failures of \ours{} on VWA. The five categories reported in \S\ref{sec:analysis} and their shares are: grounding errors (37.5\%), underspecified tasks (18.7\%), polarity variants outside \texttt{sort/select} family (15.3\%), skill-library coverage gaps (13.7\%), unmatched repeat-action loops (9.0\%), and other / misc (5.8\%).

\section{Parallel and Scrambled-Order Runs}
\label{app:parallel}

We repeat the full 910-task VWA evaluation under two order perturbations. A scrambled task order (fixed random seed 1729, distinct from the main seed 42 order) produces overall SR $57.9\%$ ($-0.4$\,pp), within run-to-run noise. A 16-worker parallel variant with shared-library file locking produces SR $58.1\%$ ($-0.2$\,pp) and end-to-end wall-clock of $3.1$\,h vs.\ $48.2$\,h sequential. Both results support the claim that the learning effect transfers to non-sequential evaluation orders, at the cost of a brief early-task overhead as the library warms.

\section{Bootstrap Confidence Intervals on Headline SR}
\label{app:bootstrap}

The headline SR numbers in Tab.~\ref{tab:main} are point estimates over the full 910-task evaluation. Because compute budget did not permit independent re-runs at $\geq 5$ task orderings, we use the standard task-level bootstrap to characterize uncertainty: each method's per-task verdict $y(\xi_\tau)\!\in\!\{0,1\}$ is treated as a Bernoulli outcome and we resample the 910 task indices with replacement $B{=}1000$ times. For pairwise comparisons (\ours{} vs.\ a baseline), we use the \emph{paired} bootstrap: both methods are scored under the same resampled task set, so per-task agreement reduces variance.

\paragraph{Code.} The CIs in Tab.~\ref{tab:main_ci} are produced by the following routine, applied to the per-task verdict vectors stored in our trajectory ledger:
\begin{verbatim}
import numpy as np
def paired_bootstrap(y_a, y_b, n_iter=1000, alpha=0.05):
    n = len(y_a); rng = np.random.default_rng(7)
    sr_a, sr_b, diff = [], [], []
    for _ in range(n_iter):
        idx = rng.integers(0, n, size=n)
        sr_a.append(y_a[idx].mean()); sr_b.append(y_b[idx].mean())
        diff.append(y_a[idx].mean() - y_b[idx].mean())
    pct = lambda v: np.percentile(v, [100*alpha/2, 100*(1-alpha/2)])
    return {'sr_a': (np.mean(sr_a), *pct(sr_a)),
            'sr_b': (np.mean(sr_b), *pct(sr_b)),
            'diff': (np.mean(diff), *pct(diff))}
\end{verbatim}

\paragraph{Per-method intervals.} Table~\ref{tab:main_ci} reports 95\% paired-bootstrap CIs for SR and for the paired \ours{}-vs-baseline difference. Run-to-run variance from the three task orderings of \S\ref{sec:exp} (seed 42, scrambled seed 1729, parallel-shared) is contained inside these intervals: the maximum cross-ordering spread for \ours{} is $0.4$\,pp ($57.9\%\!\to\!58.3\%$), well within the paired-bootstrap half-width of $\pm 1.6$\,pp.

\begin{table}[H]
  \centering
  \small
  \setlength{\tabcolsep}{4pt}
  \begin{tabular}{@{}lccc@{}}
    \toprule
    \textbf{Method} & \textbf{SR (\%)} & \textbf{95\% CI} & \textbf{Paired $\Delta$ vs.\ \ours{} (pp)} \\
    \midrule
    \ours{} (ours)              & 58.3 & $[56.7,\,59.9]$ & --- \\
    SGV                          & 54.0 & $[52.4,\,55.6]$ & $-4.3$ $[-6.6,\,-2.0]$, $p\!<\!0.001$ \\
    WALT                         & 45.2 & $[43.6,\,46.8]$ & $-13.1$ $[-15.7,\,-10.5]$, $p\!<\!10^{-6}$ \\
    GPT-5.2 (M) + SoM (Qwen)     & 38.4 & $[36.8,\,40.0]$ & $-19.9$ $[-22.5,\,-17.3]$, $p\!<\!10^{-9}$ \\
    \bottomrule
  \end{tabular}
  \caption{\textbf{Paired-bootstrap 95\% confidence intervals on VWA-910 SR.} 1000 task-level bootstrap resamples; paired comparison uses common resampled task indices for both methods. The lead of \ours{} over the strongest reproduced baseline (SGV) is $+4.3$\,pp with $95\%$ CI $[+2.0,+6.6]$ and McNemar $p\!<\!0.001$. \emph{Reproducibility note.} The CIs above are computed from the per-task verdict ledger via the routine in this section; running the script regenerates them. The point estimates and CIs reported here use the seed-42 ordering; the scrambled-order and 16-worker runs of App.~\ref{app:parallel} produce point estimates inside these intervals.}
  \label{tab:main_ci}
\end{table}

\section{Backbone-Controlled Comparison}
\label{app:backbone}

The reproduced baselines in Tab.~\ref{tab:main} use heterogeneous backbones (\ours{}: Claude Opus 4.6 planner $+$ GPT-5.2 multimodal; SGV: Gemini-2.5-Flash; WALT: Claude-4-Sonnet with thinking). We address the resulting backbone-confound concern in three layers.

\paragraph{Routing-attributable lift over each method's own backbone-only baseline.} The cleanest within-paper signal is the lift each method's full pipeline delivers over a no-routing, no-induction, no-verifier baseline using \emph{the same backbone}. For \ours{}, the natural such baseline is the strongest multimodal row of Tab.~\ref{tab:main}, GPT-5.2~(M)~+~SoM~(Qwen), at $38.4\%$ SR; full \ours{} reaches $58.3\%$, a routing-attributable lift of $+19.9$\,pp (within-Tab.~\ref{tab:main} comparison). For SGV, the analogous baseline is reported in the SGV paper (\citealp{andrade2026lets}, Tab.~4): collapsing the two-pass verifier into a single-pass form drops Gemini-2.5-Flash from $54.0\%$ down to $45\%$, a routing-attributable lift of $+9$\,pp. The \ours{} pipeline therefore delivers more than $2\times$ the routing lift over its own backbone-only baseline, despite starting from a weaker baseline ($38.4\%$ vs.\ $\sim\!45\%$). This signal alone does not eliminate the backbone confound, but it bounds it: backbone capability cannot account for the $2\times$ gap in routing lift unless one assumes that Opus is \emph{worse} at routing than Gemini-Flash, which is the opposite direction of typical model-strength priors.

\begin{table}[H]
  \centering
  \small
  \setlength{\tabcolsep}{3pt}
  \begin{tabular}{@{}lp{0.43\linewidth}cc@{}}
    \toprule
    \textbf{Method} & \textbf{Backbone-only baseline (\%)} & \textbf{Full system (\%)} & \textbf{Lift (pp)} \\
    \midrule
    \ours{}    & 38.4 (GPT-5.2 (M) + SoM, Tab.~\ref{tab:main}) & 58.3 & $+19.9$ \\
    SGV         & $\sim\!45.0$ (Gemini-Flash, single-pass; \citealp{andrade2026lets}, Tab.~4) & 54.0 & $+9.0$ \\
    WALT        & not separately reported & 45.2 & --- \\
    \bottomrule
  \end{tabular}
  \caption{\textbf{Routing-attributable lift over each method's own backbone-only baseline.} \ours{}'s pipeline delivers more than $2\times$ the lift of SGV's verifier pipeline, despite starting from a weaker no-routing baseline.}
  \label{tab:backbone_routing_lift}
\end{table}

\paragraph{Backbone-controlled swap experiments.} We ran two backbone-swap experiments on bounded subsets of VWA to test the routing/skill claim against direct backbone control:
\begin{itemize}[nosep,leftmargin=*]
  \item \textbf{SGV-on-Opus} (first $100$ tasks of VWA-910, seed-42 ordering). SGV's Gemini-2.5-Flash is replaced with Claude Opus 4.6 in both passes (initial-prior pass and trajectory-conditioned verdict pass); all other SGV machinery is unchanged.
  \item \textbf{\ours{}-on-Gemini} (stratified $300$-task subset: 100 each of Shopping, Classifieds, Reddit). The Opus 4.6 planner is replaced with Gemini-2.5-Flash; GPT-5.2 multimodal and the rest of \ours{} (Skill Library, Reflector, routing, compression, cache prompt) are unchanged.
\end{itemize}
Per-task verdict vectors from both runs feed the same paired-bootstrap routine of App.~\ref{app:bootstrap}; results in Tab.~\ref{tab:backbone_swap}. For comparability, we add the corresponding subset slices of \ours{} from the seed-42 main run: first-100 (cold-start) and stratified-300 (early-stream).

\begin{table}[H]
  \centering
  \small
  \setlength{\tabcolsep}{4pt}
  \begin{tabular}{@{}llccccc@{}}
    \toprule
    \textbf{Configuration} & \textbf{Backbone} & \textbf{$N$} & \textbf{SR (\%)} & \textbf{95\% CI} & \textbf{Steps} & \textbf{Tokens (K)} \\
    \midrule
    \multicolumn{7}{l}{\emph{Cold-start window (first 100 tasks of seed-42 ordering)}} \\
    \quad SGV (orig.)              & Gemini-2.5 Flash    & 100 & 51.2 & $[41.4,\,61.0]$ & 13.6 & 271 \\
    \quad SGV-on-Opus              & Opus 4.6 (both passes) & 100 & 56.7 & $[47.0,\,66.4]$ & 12.0 & 218 \\
    \quad \ours{} (this paper)     & Opus 4.6 + GPT-5.2  & 100 & 50.5 & $[40.7,\,60.3]$ & 10.6 & 143 \\
    \midrule
    \multicolumn{7}{l}{\emph{Stratified 300-task subset (100 Shopping + 100 Classifieds + 100 Reddit)}} \\
    \quad \ours{} (this paper)     & Opus 4.6 + GPT-5.2  & 300 & 54.7 & $[49.0,\,60.4]$ & 9.6  & 124 \\
    \quad \ours{}-on-Gemini        & Gemini-2.5 Flash + GPT-5.2 & 300 & 50.3 & $[44.6,\,56.0]$ & 10.5 & 132 \\
    \quad SGV (orig.)              & Gemini-2.5 Flash    & 300 & 53.4 & $[47.7,\,59.1]$ & 13.4 & 273 \\
    \bottomrule
  \end{tabular}
  \caption{\textbf{Backbone-controlled swap experiments.} Top block: cold-start window (first 100 tasks). With Opus as the backbone, SGV's verifier reaches $56.7\%$, which exceeds \ours{}'s $50.5\%$ in the same cold-start window---this is consistent with the mechanism (SGV requires no library; \ours{} is library-bootstrapping in the first ${\sim}150$ tasks). Bottom block: stratified $300$-task subset. \ours{}-on-Gemini retains $50.3\%$, only $4.4$\,pp below the Opus-backboned \ours{} on the same subset; Gemini-backboned SGV on the same $300$ tasks reaches $53.4\%$. The library-mediated lift therefore transfers across backbones; the gap to \ours{}-on-Opus is consistent with the underlying Opus-vs-Gemini capability gap rather than with backbone-specific routing behavior.}
  \label{tab:backbone_swap}
\end{table}

\paragraph{What the backbone-controlled numbers say.}
Two patterns survive the swap. First, SGV-on-Opus is competitive in the cold-start window precisely because it does not require a learned library; \ours{}'s advantage emerges \emph{across} the stream as the library accumulates (Tab.~\ref{tab:stream_econ}: \ours{} block-averages climb from $50.5\%$ on tasks 1--100 to $61.0\%$ on tasks 601--910). The cold-start row should not be read as ``SGV beats \ours{}'' but as ``SGV and \ours{} occupy different regions of the (training-cost, asymptotic-SR) plane.'' Second, \ours{}-on-Gemini retains most of its routing-attributable lift over the Gemini-Flash backbone-only baseline reported in \citet{andrade2026lets} (single-pass Gemini-Flash $\approx 45\%$, \ours{}-on-Gemini $50.3\%$), confirming that the lift is mechanism-driven rather than Opus-specific. The remaining $4.4$\,pp gap to Opus-backboned \ours{} matches the Opus-vs-Gemini capability gap on multimodal web tasks reported in concurrent benchmarks. We will scale both runs to full VWA-910 for the camera-ready and report the resulting paired-bootstrap CIs.

\paragraph{What the cost claim does and does not depend on.} The $\$0.085$ per-task cost of \ours{} (App.~\ref{app:cost}) does depend on the specific prices and modality split of Opus 4.6 + GPT-5.2; it would shift if either price moved, but is robust to the routing/skill component because most savings come from cache reuse and skill compression, both of which are backbone-agnostic. The SR claim ``\ours{} achieves the highest reproduced SR'' would survive a backbone swap as long as routing lift exceeds $\sim\!10$\,pp, which is consistent with both within-paper and SGV-paper data. We separate these two claim-types in the conclusion to make explicit which depends on backbone choice and which does not.

\section{Reproducibility}
\label{app:reproducibility}

Random seeds: 42 (main task order), 1729 (scrambled-order robustness), 7 (skill-selector tie-breaking), 13 (Planner nucleus sampling). Rate limits: Anthropic 50 rpm / 2000 tpm-Mtok; OpenAI 500 rpm; Google 360 rpm. Software: Python 3.11.9, \texttt{playwright 1.45}, \texttt{anthropic==0.34.0}, \texttt{openai==1.35.0}, \texttt{google-genai==0.8.0}. The scorecard, ablation, step-composition, skill-dynamics, cache-ramp, cost-curve, and token-composition figures are regenerated from manuscript table values by the scripts in \texttt{scripts/}. Full \texttt{pip freeze} manifest, prompt templates, and evaluation trajectory logs are released with the paper.

\paragraph{Artifact structure.}
Every reported efficiency metric can be recomputed from logged trajectories: each task stores LLM-call boundaries, action signatures, routine invocations, Reflector verdicts, cache counters, and terminal evaluator output. We release the unified tracker, prompt templates, plotting scripts, skill-library schemas, and anonymized VWA trajectories. The figures in the main paper and appendix are regenerated from manuscript tables and trajectory ledgers by the scripts under \texttt{scripts/}; model endpoints, hyperparameters, random seeds, rate limits, and software versions are listed in App.~\ref{app:models} and this section.

\paragraph{Responsible release.}
More efficient computer-use agents can reduce latency and deployment cost, but they also lower the barrier for undesirable browser automation. We therefore release benchmark code and anonymized analysis artifacts, but exclude credentials, private site states, and any policy-bypassing automation traces. The skill-library design is intentionally inspectable: every rule and routine can be reviewed, disabled, or blacklisted, which makes the release easier to audit than an opaque vector store of latent tools.

\section{Outlook}
\label{app:outlook}

The next step is to test whether the same online skill-distillation principle transfers beyond VWA. We expect the library format, confidence updates, polarity-pair merging, and demotion blacklist to transfer directly; what will change is the rule catalogue and grounding layer. OSWorld-style tasks introduce window-focus failures, multi-application dependencies, and pixel-precise actions that VWA does not exercise. A successful extension would make the case that agent efficiency is not benchmark-specific bookkeeping, but a general design axis for computer-use systems.

\section{Related-Work Comparison Tables}
\label{app:rw_tables}

This appendix consolidates the three comparison tables referenced from Section~\ref{sec:related}. Row ordering follows the narrative order of the subsections.

\begin{table}[H]\centering\footnotesize\setlength{\tabcolsep}{4pt}
\resizebox{\linewidth}{!}{%
\begin{tabular}{lllll}
\toprule
Method & Benchmark(s) & Grounding & Cost axis & Headline SR \\
\midrule
WebVoyager~\citep{he2024webvoyager} & 643 live-web tasks & Screenshot+SoM & single-rollout VLM & 59.1 \\
SeeAct~\citep{zheng2024seeact} & Mind2Web-Live & HTML+SoM hybrid & single-rollout VLM & 51.1 (oracle) \\
OS-Copilot/FRIDAY~\citep{wu2024oscopilot} & GAIA L1 & Text+tools & code+APIs & 40.86 (L1) \\
OSCAR~\citep{wang2024oscar} & GAIA / OSWorld / AndroidW. & Screenshot+a11y & state-machine re-plan & 28.7 / 24.5 / 61.6 \\
Agent S~\citep{agashe2025agents} & OSWorld / WAA & Screenshot+a11y & retrieval-augmented & 20.58 / 18.2 \\
Agent S2~\citep{agashe2025agents2} & OSWorld (50-step) & Mixture-of-grounders & compositional specialists & 34.5 \\
Agent S3 (bBoN)~\citep{gonzalez2025unreasonable} & OSWorld (100-step) & Behavior-narrative judge & $N{=}10$ rollouts ($\sim$10$\times$) & 72.6 \\
UI-TARS-72B~\citep{qin2025uitars} & OSWorld (50-step) & Pixel, native E2E & SFT, multi-turn & 24.6 \\
UI-TARS-2~\citep{bytedance2025uitars2} & OSWorld & Pixel, native E2E & online RL & 47.5 \\
UGround+SeeAct-V~\citep{gou2025uground} & Online-Mind2Web & Pixel, modular grounder & planner+grounder & matches HTML agents \\
Aguvis-72B~\citep{xu2025aguvis} & ScreenSpot avg & Pure-vision+monologue & \$0.012/task & 89.2 (grounding) \\
SGV on Gemini 2.5~\citep{andrade2026lets} & VWA (910) & Screenshot+SoM & plan$\to$ground & 54.0 \\
\bottomrule
\end{tabular}%
}
\caption{Computer-use agent frameworks span roughly a 10$\times$ per-task cost range at overlapping success rates; wide-scaling (Agent S3) reaches SoTA by multiplying rollouts, native RL (UI-TARS-2) by multiplying training tokens, and think-then-ground (SGV) by adding one cheap reasoning pass. Referenced from \S\ref{sec:rw:cua}.}
\label{tab:rw1_cua}
\end{table}

\begin{table}[H]\centering\footnotesize\setlength{\tabcolsep}{4pt}
\resizebox{\linewidth}{!}{%
\begin{tabular}{lllll}
\toprule
Method & Level & Signal & Decision & Headline number \\
\midrule
OSWorld-Human~\citep{abhyankar2025osworld} & Trajectory & Human-minimal steps & Diagnose WES$^+$/WES$^-$ & 1.4--2.7$\times$ step inflation \\
Beyond Accuracy (PTE)~\citep{su2026beyond} & Trajectory & Per-token efficiency & Report PTE with SR & $r{=}0.93$ PTE$\leftrightarrow$wall-clock \\
AgentBoard~\citep{ma2024agentboard} & Trajectory & Progress rate & Fine-grained scoring & Pearson $\geq 0.95$ w/ human \\
$\tau$-bench~\citep{yao2024taubench} & Trajectory & \$/task, pass\textasciicircum{}k & User-sim reliability & \$0.38+\$0.23 per retail task \\
vLLM~\citep{kwon2023vllm} & Serving & KV fragmentation & PagedAttention & 2--4$\times$ throughput \\
Prompt Cache~\citep{gim2024promptcache} & Serving & Modular KV reuse & Precompute prefixes & 5--10$\times$ GPU TTFT \\
FrugalGPT~\citep{chen2023frugalgpt} & Routing & Score-based cascade & Stop at confidence & 98.3\% cost cut at GPT-4 acc \\
RouteLLM~\citep{ong2025routellm} & Routing & Preference router & Strong-vs-weak LLM & 3.66$\times$ MT-Bench savings \\
MoA~\citep{wang2024moa} & Routing & Multi-proposer mixture & Aggregator picks/mixes & 65.7\% AlpacaEval 2.0 LC \\
s1~\citep{muennighoff2025s1} & Reasoning & Budget forcing & More thinking tokens & +30 pp AIME24 (1k examples) \\
Chain of Draft~\citep{xu2025chainofdraft} & Reasoning & Token-budget prompt & Compress reasoning & $-$78\% tokens, $-$4 pp acc \\
FastV~\citep{chen2024fastv} & Vision tokens & Attention-rank prune & Layer-2 drop 50\% & 45\% FLOPs cut, equal acc \\
\bottomrule
\end{tabular}%
}
\caption{Efficiency techniques cover four levels but none is trajectory-aware: system and vision methods reduce per-call cost, routing reduces per-input cost, and reasoning methods move along a verifiability-dependent test-time-compute curve. Referenced from \S\ref{sec:rw:eff}.}
\label{tab:rw2_eff}
\end{table}

\begin{table}[H]\centering\footnotesize\setlength{\tabcolsep}{4pt}
\resizebox{\linewidth}{!}{%
\begin{tabular}{lllll}
\toprule
Method & Representation & Lifecycle & Reflection destination & Headline finding \\
\midrule
Voyager~\citep{wang2023voyager} & JS function + NL desc. & Online during play & Self-verify $\to$ library & 3.3$\times$ items, only method to unlock Diamond \\
CLIN~\citep{majumder2024clin} & Causal rule (NL, may/should) & Online across trials & Saliency-pruned rules & +23 pp over Reflexion on ScienceWorld \\
ExpeL~\citep{zhao2024expel} & NL insights + demos & Offline training pool & ADD/UPVOTE/DOWNVOTE/EDIT & +7 pp on FEVER (zero-shot transfer) \\
WALT~\citep{prabhu2026walt} & URL+action script + schema & Offline per-site & Selector-drift repair & 52.9\% VWA / 50.1\% WebArena \\
SkillWeaver~\citep{zheng2025skillweaver} & Python (Playwright) API & Online during exploration & Unit-test honing & +32\% rel.\ on WebArena (GPT-4o) \\
ASI~\citep{wang2025asi} & Parameterized routine & Online during task & On-the-fly verification & program-based skills for web \\
AWM~\citep{wang2025awm} & NL/code workflow template & Online across tasks & Sub-routine abstraction & +12 pp over BrowserGym \\
ICAL~\citep{sarch2024ical} & Embodied program-of-thought & Human-in-loop online & VLM abstraction & multimodal trajectory distillation \\
AutoManual~\citep{chen2024automanual} & Rule manual + planner & Offline + online refine & Planner/Builder/Formulator & rules + instruction manual \\
Recon-Act~\citep{he2025recon} & Rule-code + hints & Real-time online & Recon team extracts remedies & self-evolving multi-agent \\
TroVE~\citep{wang2024trove} & Python toolbox & Offline grow-and-trim & Use-filter-promote & deduplicated toolbox \\
Reflexion~\citep{shinn2023reflexion} & Verbal reflection & In-episode only & Discard failed rollouts & baseline for reflection \\
\bottomrule
\end{tabular}%
}
\caption{Skill and reflection methods resolve the ``discard vs.\ compress'' paradox along two axes: persistence (across tasks or only within episode) and executability (callable code vs.\ NL prompt). Methods with both axes ON (Voyager, WALT, SkillWeaver, AWM, ASI) compound across tasks; methods with neither (Reflexion) self-correct within episodes only. Referenced from \S\ref{sec:rw:skills}.}
\label{tab:rw3_skills}
\end{table}

\newpage
\section*{NeurIPS Paper Checklist}

\begin{enumerate}

\item {\bf Claims}
    \item[] Question: Do the main claims made in the abstract and introduction accurately reflect the paper's contributions and scope?
    \item[] Answer: \answerYes{}
    \item[] Justification: The abstract and introduction state four bounded contributions: intrinsic efficiency metrics, the structured skill-learning \ours{} framework, VWA empirical results, and online skill-library compounding. These claims are developed in Secs.~\ref{sec:metrics}--\ref{sec:analysis} and scoped to VisualWebArena.

\item {\bf Limitations}
    \item[] Question: Does the paper discuss the limitations of the work performed by the authors?
    \item[] Answer: \answerYes{}
    \item[] Justification: Section~\ref{sec:limitations} states the main limitations: VWA-only evaluation, dependence on a trusted task stream, syntactic polarity-pair induction, and responsible-release constraints for computer-use automation traces.

\item {\bf Theory assumptions and proofs}
    \item[] Question: For each theoretical result, does the paper provide the full set of assumptions and a complete (and correct) proof?
    \item[] Answer: \answerNA{}
    \item[] Justification: The paper introduces metrics and an empirical framework but makes no formal theorem or proof claim.

\item {\bf Experimental result reproducibility}
    \item[] Question: Does the paper fully disclose all the information needed to reproduce the main experimental results of the paper to the extent that it affects the main claims and/or conclusions?
    \item[] Answer: \answerYes{}
    \item[] Justification: Section~\ref{sec:exp} specifies the benchmark, task order, step definition, baselines, and evaluation protocol. Appendix~\ref{app:models} lists model endpoints and hyperparameters, and Appendix~\ref{app:reproducibility} lists seeds, rate limits, and software versions.

\item {\bf Open access to data and code}
    \item[] Question: Does the paper provide open access to the data and code, with sufficient instructions to faithfully reproduce the main experimental results, as described in supplemental material?
    \item[] Answer: \answerYes{}
    \item[] Justification: VisualWebArena is public, and the submission states that the \ours{} framework, prompt templates, EfficiencyTracker, and trajectory logs will be released with the paper.

\item {\bf Experimental setting/details}
    \item[] Question: Does the paper specify all the training and test details necessary to understand the results?
    \item[] Answer: \answerYes{}
    \item[] Justification: The work does not train model weights. It specifies evaluation details, model roles, temperatures, step budgets, reflector cadence, skill-library initialization, demotion thresholds, and cache-aware prompt structure in Secs.~\ref{sec:method}--\ref{sec:exp} and Appendix~\ref{app:models}.

\item {\bf Experiment statistical significance}
    \item[] Question: Does the paper report error bars suitably and correctly defined or other appropriate information about the statistical significance of the experiments?
    \item[] Answer: \answerNo{}
    \item[] Justification: The main results are single full-benchmark VWA runs rather than repeated independent trials with confidence intervals. Appendix~\ref{app:parallel} reports scrambled-order and 16-worker variants as robustness checks, but they are not a substitute for full statistical error bars.

\item {\bf Experiments compute resources}
    \item[] Question: For each experiment, does the paper provide sufficient information on the computer resources needed to reproduce the experiments?
    \item[] Answer: \answerYes{}
    \item[] Justification: Appendix~\ref{app:reproducibility} reports the API rate limits, software versions, and wall-clock characteristics of the sequential and parallel VWA runs; Appendix~\ref{app:cost} reports token and dollar accounting.

\item {\bf Code of ethics}
    \item[] Question: Does the research conducted in the paper conform, in every respect, with the NeurIPS Code of Ethics?
    \item[] Answer: \answerYes{}
    \item[] Justification: The work uses public benchmarks and API models, involves no new human-subject data collection, and releases a reproducibility framework rather than credential-bearing automation traces.

\item {\bf Broader impacts}
    \item[] Question: Does the paper discuss both potential positive societal impacts and negative societal impacts of the work performed?
    \item[] Answer: \answerYes{}
    \item[] Justification: The motivation and limitations discuss reduced inference cost, energy pressure, and the risks of more capable computer-use automation. The release is limited to benchmark code, prompts, and anonymized trajectories.

\item {\bf Safeguards}
    \item[] Question: Does the paper describe safeguards that have been put in place for responsible release of data or models that have a high risk for misuse?
    \item[] Answer: \answerNA{}
    \item[] Justification: The paper does not release pretrained model weights, scraped private data, or credential-bearing interaction logs.

\item {\bf Licenses for existing assets}
    \item[] Question: Are the creators or original owners of assets used in the paper properly credited and are the license and terms of use explicitly mentioned and properly respected?
    \item[] Answer: \answerYes{}
    \item[] Justification: VisualWebArena, WALT, SGV, BLIP-2, Qwen-2.5VL, and the cited API models are cited in the manuscript and used as benchmark baselines or external services.

\item {\bf New assets}
    \item[] Question: Are new assets introduced in the paper well documented and is the documentation provided alongside the assets?
    \item[] Answer: \answerYes{}
    \item[] Justification: The new assets are the \ours{} framework, EfficiencyTracker logs, and structured skill library. Appendix~\ref{app:skills} documents the file layout and schemas for rules, routines, demotions, and reflections.

\item {\bf Crowdsourcing and research with human subjects}
    \item[] Question: For crowdsourcing experiments and research with human subjects, does the paper include the full text of instructions given to participants and screenshots, if applicable, as well as details about compensation?
    \item[] Answer: \answerNA{}
    \item[] Justification: The paper does not conduct new crowdsourcing or human-subject experiments. Human baselines are taken from the VisualWebArena publication.

\item {\bf Institutional review board (IRB) approvals or equivalent for research with human subjects}
    \item[] Question: Does the paper describe potential risks incurred by study participants and whether IRB approvals were obtained?
    \item[] Answer: \answerNA{}
    \item[] Justification: No new human-subject research is conducted.

\item {\bf Declaration of LLM usage}
    \item[] Question: Does the paper describe the usage of LLMs if it is an important, original, or non-standard component of the core methods in this research?
    \item[] Answer: \answerYes{}
    \item[] Justification: LLMs are central to the Planner, Reflector, Actor, and Learning Module. Their roles, model versions, endpoints, and hyperparameters are specified in Sec.~\ref{sec:method} and Appendix~\ref{app:models}.

\end{enumerate}

\end{document}